\documentclass[journal]{IEEEtran}
\ifCLASSINFOpdf
\else
\fi
\usepackage{amsmath}\usepackage{epsfig}
\usepackage{graphicx}
\usepackage{amsmath}

\usepackage{amssymb}
\usepackage{algorithm}
\usepackage{algpseudocode}
\usepackage{mathrsfs}
\usepackage{amssymb}
\usepackage{amsthm}\usepackage{amssymb}
\usepackage{amsthm}
\newtheorem{theorem}{Theorem}

\newtheorem{lemma}[theorem]{Lemma}

\theoremstyle{definition}

\newtheorem{Problem Statement}{Problem Statement}

\begin{document}
%
\title{Analysis Dictionary Learning based Classification: Structure for Robustness}
%
%
%

\author{Wen~Tang,~\IEEEmembership{Member,~IEEE,}
        Ashkan~Panahi,~\IEEEmembership{Member,~IEEE,}
        Hamid~Krim,~\IEEEmembership{Fellow,~IEEE,}
        and Liyi~Dai,~\IEEEmembership{Fellow,~IEEE}
\thanks{W. Tang, A. Panahi and H. Krim are with the Department
of Electrical and Computer Engineering, North Carolina State University, Raleigh, NC 27606 USA.}
\thanks{Email: wtang6@ncsu.edu; apanahi@ncsu.edu; ahk@ncsu.edu}
\thanks{L. Dai is with Raytheon Integrated Defense Systems, Tewksbury, MA 01876 USA.}
\thanks{Email: liyi.dai@raytheon.com}
\thanks{This paper has been accepted and published by IEEE Transactions on Image Processing. The Appendix part in this paper is the supplementary material in published version.}
}

\maketitle

\begin{abstract}
A discriminative structured analysis dictionary is proposed for the classification task. A structure of the union of subspaces (UoS) is integrated into the conventional analysis dictionary learning to enhance the capability of discrimination. A simple classifier is also simultaneously included into the formulated function to ensure a more complete consistent classification. The solution of the algorithm is efficiently obtained by the linearized alternating direction method of multipliers. Moreover, a distributed structured analysis dictionary learning is also presented to address large scale datasets. It can group-(class-) independently train the structured analysis dictionaries by different machines/cores/threads, and therefore avoid a high computational cost. A consensus structured analysis dictionary and a global classifier are jointly learned in the distributed approach to safeguard the discriminative power and the efficiency of classification. Experiments demonstrate that our method achieves a comparable or better performance than the state-of-the-art algorithms in a variety of visual classification tasks. In addition, the training and testing computational complexity are also greatly reduced.
\end{abstract}

\begin{IEEEkeywords}
Discriminate analysis dictionary learning, distributed analysis dictionary learning, structured mapping, supervised learning.
\end{IEEEkeywords}

%
\IEEEpeerreviewmaketitle

\section{Introduction}
%
%
%
%
\IEEEPARstart{S}{parse} representation has had great success in dealing with various problems in image processing and computer vision, such as image denoising and image restoration. To obtain such sparse representations with an unknown precise model, Dictionary Learning is one choice because it results in a linear combination of sparse dictionary atoms. There are two different types of dictionary learning methods: Synthesis Dictionary Learning (SDL) and Analysis Dictionary Learning (ADL).

\begin{figure}[htb]
	\centering
	\includegraphics[width=0.40\textwidth]{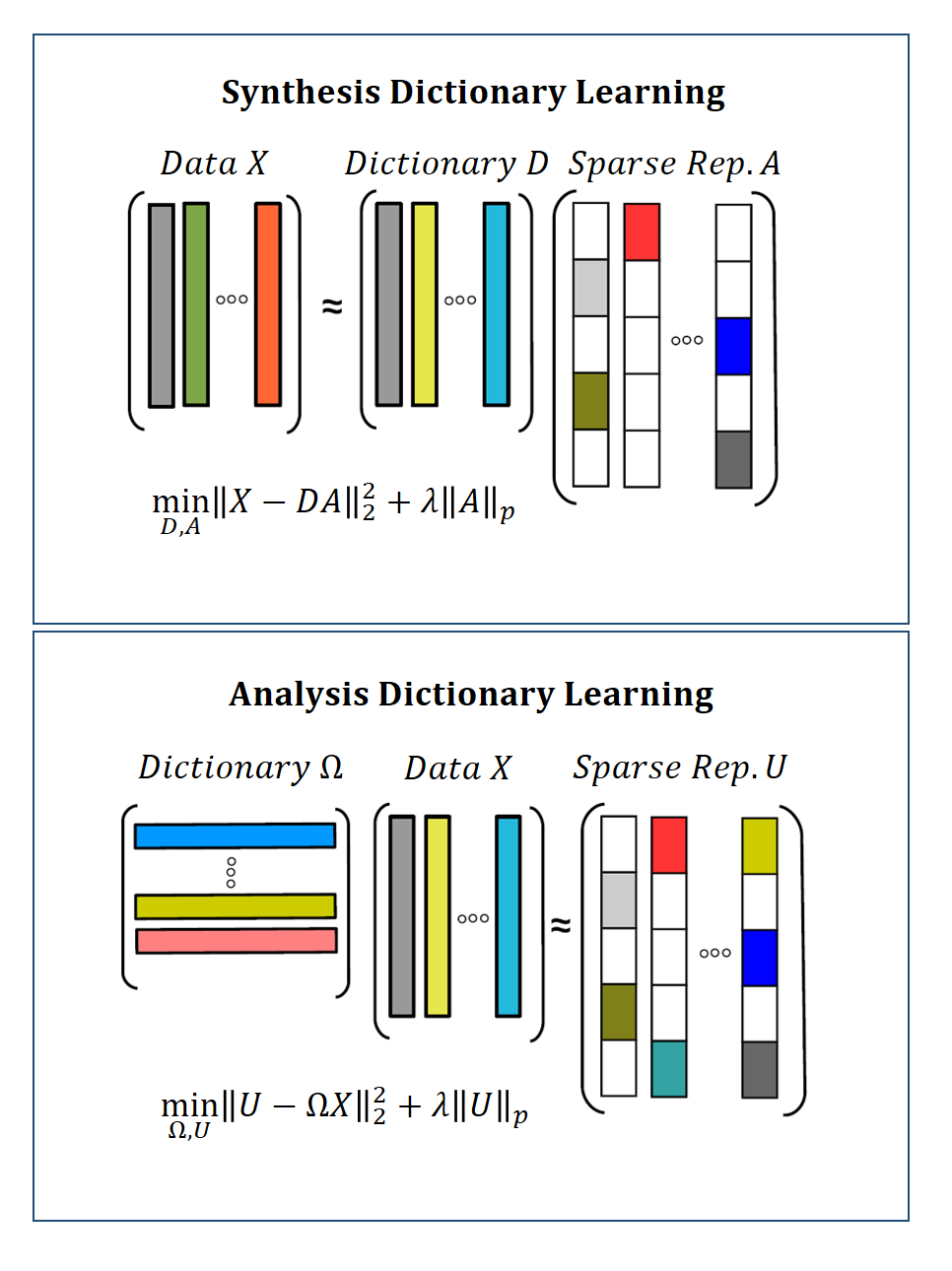}
	\caption{SDL reconstructs data $X$ by the dictionary $D$ with the sparse representations $A$. ADL applies the dictionary $\Omega$ to data $X$ and results in the sparse representations $U$. $\| \cdot \|_p$ can be either $l_1$ norm or $l_0$ norm. If and only if $D$ and $\Omega$ are square matrices, SDL and ADL are equivalent to each other.}
	\label{fig:sdlvsadl}
\end{figure}

In recent years, SDL has been prevalently and widely studied \cite{olshausen1996emergence,Bruck09,mahdizadehaghdam2018deep}, while ADL has received little attention. SDL supposes that a signal lies in a sparse latent subspace and can be recovered by an associated dictionary. The local structures of the signal are well preserved in the optimal synthesis dictionary \cite{mairal2008sparse,mairal2009non,mairal2009online}. In contrast, ADL assumes that a signal can be transformed into a latent sparse subspace by its corresponding dictionary. In other words, ADL is to produce a sparse representation by applying the dictionary as a transform to a signal. The atoms in an analysis dictionary can be interpreted as local filters, as first mentioned in \cite{roth2009fields}. Sparse representations can be simply obtained by an inner product operation, when the dictionary is known. Such a fast coding supports ADL more favored than SDL in applications. The contrast of SDL and ADL is shown in Fig. \ref{fig:sdlvsadl}.\par

The success of dictionary learning in image processing problems has shaped much interest in task-driven dictionary learning methods for inference applications, such as image classification. The task of classification aims to assign the correct label to an observed image, which requires a much more discriminative capacity of either the dictionary or the sparse representation. Towards addressing this issue, supervised learning is often invoked when using single-view SDL \cite{mairal2009supervised} so as to maximize the distances between the sparse representations of each of two distinct classes. In addition, multi-view dictionary learning methods \cite{liu2019online,jing2014uncorrelated,bahrampour2016multimodal,wang2016semi} were developed to include more information of each class. 

For the supervised single-view dictionary learning methods, there are generally two strategies to address the supervised learning approaches. The first strategy is to learn multiple dictionaries or class-specific dictionaries for different classes \cite{src,ramirez2010classification,yang2011fisher,wang2013max}. The advantage of learning multiple dictionaries is that these dictionaries characterize specific patterns and structures of each class and enhance the distances between different classes. The minimum reconstruction errors of various dictionaries are subsequently used to assign labels of new incoming images. In \cite{ramirez2010classification}, Ramirez \emph{et al.} learned class-specific dictionaries with penalty for the common atoms. Yang \emph{et al.} \cite{yang2011fisher} then learned class-specific dictionaries and jointly applied a Fisher criterion to  associative sparse representations to thereby enhance the distances between each class. A large-margin method was proposed to increase the divergence of sparse representations for the class-specific dictionaries in \cite{wang2013max}. However, as the number of classes increases, it would be too complex and time consuming to train class-specific dictionaries with regularizing distances of each dictionary. Even though a distributed cluster could reduce the time complexity of training dictionaries, it is difficult for the distributed algorithm to communicate with each independent cluster and to compromise with other regularizations for the class-specific dictionary learning.

Another strategy is to learn a shared dictionary for all classes together with a universal classifier \cite{mairal2009supervised,lcksvd}. Such a joint dictionary learning enforces more discriminative sparse representations. Compared with class-specific dictionary learning, using this strategy is simpler to learn such a dictionary and classifier, and easier to test the unknown images. In \cite{mairal2009supervised}, Mairal \emph{et al.} integrated a linear classifier in a sparse representation for a dictionary learning phase. Jiang \emph{et al.} then included a linear classifier and a label consistent regularization term to enforce more consistent sparse representations in each class \cite{lcksvd}. When any large data sets are on hand, memory and computational limitations emerge, and an online learning or distributed solutions are required as a viable strategy.

Although the techniques mentioned above are all based on SDL, ADL has gradually received more attention \cite{nam2013cosparse}. Based on the seminal work of Rubinstein \emph{et al.}  \cite{Rubi13} proposing analysis K-SVD to learn an analysis dictionary, Li \emph{et al.}  \cite{li2014dictionary} considered to learn ADL by using an additional inner product term of sparse coefficients to increase its discriminative power. In addition, reducing the computational complexity has been addressed in recent methods. Zhang \emph{et al.} \cite{zhang2014analysis} use Recursive Least Square method to accelerate dictionary learning by updating a dictionary based on the dictionaries in the previous iterations. Li \emph{et al.} \cite{li2015analysis} and Dong \emph{et al.} \cite{dong2016analysis} proposed Simultaneous codeword optimization (SimCo) related algorithms to update multiple atoms in each iteration and by adding an extra incoherent regularity term to avoid linear dependency among dictionary atoms. On other hand, Li \emph{et al.} \cite{li2016constrained, li2018proximal} used $l_{1/2}$ norm instead of $l_1$ norm to have stronger sparsity and mathematically guaranteed a strong convergence. In \cite{bian2016sparsity}, Bian \emph{et al.} \cite{bian2016sparsity} proposed the Sparse Null Space (SNS) pursuit to search for the optimal analysis dictionary. However, all of these methods are proposed for the original problem of learning an analysis dictionary. To the best of our knowledge, few attempts have been carried out for task-driven ADL. For example, in \cite{shekhar2014analysis}, Shekhar \emph{et al.} \cite{shekhar2014analysis} learned an analysis dictionary and subsequently trained SVM for the digital and face recognition tasks. Their results demonstrate that ADL is more stable than SDL under noise and occlusion, and achieves a competitive performance. Guo \emph{et al.} \cite{dadl} integrated local topological structures and discriminative sparse labels into the ADL and separately classified images by a $k$ Nearest Neighbor classifier. Instead of preforming ADL and SVM in separate steps, Wang \cite{cadl} alternately optimize ADL and SVM together to classify different patterns. In \cite{sksvdadl}, Wang \emph{et al.} use the K-SVD based technique to solve a joint learning of ADL and a linear classifier. Additionally, a hybrid design based on both SDL and ADL is considered in \cite{DPL} and \cite{zhang2017jointly}. A multi-view ADL was proposed in \cite{wang2018multi}, which separately learns analysis dictionaries for different views and a marginalized classifier for fusing the semantic information of each view.


Inspired by these past works, and taking advantage of efficient coding by ADL, we propose a supervised ADL with a shared dictionary and a universal classifier. In addition to the classifier, a structured subspace regularization is also included into an ADL model to obtain a more structured discriminative and efficient approach to image classification. We refer to this approach as Structured Analysis Dictionary Learning (SADL).
Since Sparse Subspace Clustering\cite{elhamifar2013sparse} has shown that visual data in a class or category can be well captured and localized by a low dimensional subspace, and the sparse representation of the data within a class similarly share a low dimensional subspace, a structured representation is introduced to achieve a distinct representation of each class. This achieves more coherence for within-class sparse representations and more disparity for between-class representations. When sorted by the order of classes, these representations as shown later can be viewed as a block-diagonal matrix. For robustness of the sought sparse representations, we simultaneously learn a one-against-all regression-based classifier. The resulting optimization function is solved by a linearized alternative direction method (ADM)\cite{lin2011linearized}. This approach leads to a more computationally efficient solution than that of analysis K-SVD \cite{Rubi13} and of SNS pursuit \cite{bian2016sparsity}.
Additionally, a great advantage of our algorithm is its extremely short on-line encoding and classification time for an incoming observed image. It is easy to understand that in contrast to the SDL encoding procedure, ADL obtains a sparse representation by a simple matrix multiplication of the learned dictionary and testing data. 
Experiments demonstrate that our method achieves an overall better performance than the synthesis dictionary approach. A good accuracy is achieved in the scene and object classification with a simple classifier, and at a remarkably low computational complexity to seek the best performances of facial recognition problems. Moreover, experiments also show that our approach has a more stable performance than that of SDL. Even when the dictionary size is reduced to result in memory demand reduction, our performance is still outstanding. To address large datasets, a distributed structured analysis dictionary learning algorithm is also developed while preserving the same properties as those of structured analysis dictionary learning (SADL). Experiments also show that when the dataset is sufficient, a distributed algorithm achieves as high a performance as SADL.

The following represent our main contributions,
\begin{itemize}
	\item Both a structured representation and a classification error regularization term are introduced to the conventional ADL formulation to improve classification results. A multiclass classifier and an analysis dictionary are jointly learned.
	\item The optimal solution provided by the linearized ADM is significantly faster than other existing techniques for non-convex and non-smooth optimization.
	\item An extremely short classification time is offered by our algorithms, as they entail encoding by a mere matrix multiplication for a simple classification procedure.
	\item A distributed structured analysis dictionary learning algorithm is also presented.
\end{itemize}
\par
The balance of this paper is organized as follows: we state and formulate the problem of SADL and its distributed form in Section \ref{sec:SSADL}. The resulting solutions to the optimization problems along with the classification procedure are described in Sections \ref{sec:solve} and \ref{sec:classifier}. In Section \ref{sec:Convergeandcomplexity}, we analyze the convergence and complexity of our methods. The experimental comprehensively validation and results are then presented in Section \ref{sec:experiments}. Some comments and future works are finally provided in Section \ref{sec:conclusion1}.

\section{Structured Analysis Dictionary Learning}
\label{sec:SSADL}

\subsection{Notation}
Uppercase and lowercase letters respectively denote matrices and vectors throughout the paper. The transpose and inverse of matrices are represented as the superscripts $T$ and $-1$, such as $A^T$ and $A^{-1}$. The identity matrix and all-zero matrix are respectively denoted as $I$ and $\textbf{0}$. $(a_i)_j$ represents the $j$th element in the $i$th column of matrix $A$.
\subsection{Structured ADL Method}
\label{subsec:ssadl}
\subsubsection{ADL Formulation}
The conventional ADL problem \cite{Rubi13} aims at obtaining a representation frame $\Omega$ with a sparse coefficient set $U$ based on the data matrix $X=[x_1,\dots,x_n] \in \mathbb{R}^{m \times n}$.
\begin{equation}\label{equ:adl}
\begin{split}
\arg\min_{\Omega,U} & ~ \frac{1}{2}\|U-\Omega X\|_2^2+\lambda_1 \|U\|_1\\
s.t. &~ \Omega \in \mathbb{R}^{r\times m} \subset \mathcal{W},
\end{split}
\end{equation}
where $U \in \mathbb{R}^{r \times n}$ and $\mathcal{W}$ is a large class of non-trivial solutions.

\subsubsection{Mitigating Inter-Class Feature Interference}
The basic idea of our algorithm is to take advantage of the stability to perturbations and of the fast encoding of ADL. Since there is no reconstruction term in the conventional ADL, and to secure an efficient classification, the representation $U$ is used to obtain a classifier in a supervised learning mode. To strengthen the discriminative power of ADL, it is desirable to minimize the impacts of inter-class common features. We therefore propose two additional constraints on $U$ by way of:
\begin{itemize}
	\item Minimizing interference of inter-class common features by a structural map of $U$.
	\item Minimizing the classification error.
\end{itemize}
\paragraph{Structural Mapping of U}
The first constraint is to particularly ensure that the representation of each sample in the same class belong to a subspace defined by a span of the associated coefficients. This imposes the distinction among the classes and improves the identification of each class, and efficiently enhances the divergence between classes. Specifically, we introduce a block-diagonal matrix $H \in \mathbb{R}^{s \times n}$ as shown below,
\[H=
\bordermatrix{~& h^1_1 & h^1_2 & h^1_3 & h^2_4 & h^2_5 & h^3_6 & h^3_{7}\cr
	~& 1 & 1 & 1 & 0 & 0 & 0 & 0 \cr
	~& 1 & 1 & 1 & 0 & 0 & 0 & 0 \cr
	~& 1 & 1 & 1 & 0 & 0 & 0 & 0 \cr
	~& 0 & 0 & 0 & 1 & 1 & 0 & 0 \cr
	~& 0 & 0 & 0 & 1 & 1 & 0 & 0 \cr
	~& 0 & 0 & 0 & 0 & 0 & 1 & 1 \cr
	~& 0 & 0 & 0 & 0 & 0 & 1 & 1 \cr
}
,\]
where $s\geq n$ is the length of the structured representation. Each diagonal block in $H$ represents a subspace of each class to force each one class to remain distinct from another with a consistent intra-class representation. Each column $h_i^j$ is a structured representation for the corresponding data point, which is pre-defined on the training labels. $H$ is not necessarily a uniformly block-diagonal matrix, and the order of samples is not important, so long as the structured representation corresponds to a given class. To mildly relax the constraint, and integrate it into the ADL function, we write 
\begin{equation}
\label{equ:ssp}
H=QU+\varepsilon_1,
\end{equation}
where $Q \in \mathbb{R}^{s \times r} $ is a matrix to be learned with $\Omega$ and $U$, $\varepsilon_1$ is the tolerance. 


\paragraph{Minimal Classification Error}
To maintain an audit track on the desired representation, we include a classification error to make the representation $QU$ discriminative and learn an optimal regularization. This is written as
\begin{equation}
\label{equ:lc}	
Y=W(QU)+\varepsilon_2,
\end{equation}
where $\varepsilon_2$ is the tolerance, $W \in \mathbb{R}^{c \times s}$ is a linear transform, and the label matrix $Y \in \mathbb{R}^{c \times n}$ is defined as
$$Y_{ij}=
\begin{cases}
1& \text{if image } j \text{ belongs to class } i\\
0& \text{otherwise}
\end{cases},$$
and $c$ is the number of classes.
\par

\subsubsection{Structured ADL Formulation}
To account for all these constraints and to avoid overfitting by $l_2$ regularization arising $\Omega,Q$ and $W$, we can rewrite the ADL optimization problem as
\begin{equation}\label{equ:ssadl}
\begin{split}
\arg \min_{\substack{\Omega, U, Q, W, \\ \varepsilon_1, \varepsilon_2}} & \frac{1}{2}\|U - \Omega X\|_F^2+ \lambda_1  \|U\|_1\\
& +\frac{\rho_1}{2}\|\varepsilon_1\|^2_2+\frac{\rho_2}{2}\|\varepsilon_2\|^2_2\\ 
& +\frac{\delta_1}{2}\|Q\|^2_2+\frac{\delta_2}{2}\|W\|^2_2+\frac{\lambda_2}{2}\|\Omega\|^2_2\\
\emph{s.t.} ~&H=QU+\varepsilon_1,\\
& Y=W(QU)+\varepsilon_2,\\
\end{split}
\end{equation}
where 
$\rho_1$ and $\rho_2$ are the penalty coefficients, $\delta_1, \delta_2, \lambda_1$ and $\lambda_2$ are tuning parameters. Recall $H$ is the structured representation, $Q$ is the structuring transformation, $Y$ is the classifier label, and $W$ is the linear classifier.

The formulated optimization function in Eq. (\ref{equ:ssadl}) provides an analysis dictionary driven by the latent structure of the data yielding an improved discriminative sparse representation among numerous classes.

\subsubsection{Distributed Structured ADL Formulation}
In order to handle large datasets, we propose a distributed Structured ADL method. Since both the discriminative structure and the efficient classification need to be preserved, we introduce a global analysis dictionary, a global structuring transformation and a global classifier. In pursuing a distributed ADL, we ensure that the global variables share information with each distributed dictionary cluster, thereby ensuring that the global analysis dictionary, the structured transform and the classifier respectively reach a consensus,
\begin{equation}
\|\Omega-\Omega_t\|^2, \|Q-Q_t\|^2, \|W-W_t\|^2, \forall t=1,\dots,N.
\end{equation}
Together with the consensus penalties, the distributed SADL is formulated as
\begin{equation}
\begin{split}
\arg &\min_{\substack{\Omega_t, U_t,\\ Q_t, W_t, \\ \Omega, Q, W,\\ \varepsilon_{1_t}, \varepsilon_{2_t}}} \sum_{t=1}^N (\frac{1}{2}\|U_t - \Omega_t X_t\|_F^2+ \lambda_1  \|U_t\|_1 +\frac{\rho_{1_t}}{2}\|\varepsilon_{1_t}\|^2_2 \\
&+\frac{\rho_{2_t}}{2}\|\varepsilon_{2_t}\|^2_2 +\frac{\xi_{1_t}}{2}\|\Omega-\Omega_t\|^2_2 +\frac{\delta_{1_t}}{2}\|Q_t\|^2_2+\frac{\lambda_{2_t}}{2}\|\Omega_t\|^2_2\\
&+\frac{\xi_{2_t}}{2} \|Q-Q_t\|^2_2 +\frac{\delta_{2_t}}{2}\|W_t\|^2_2+\frac{\xi_{3_t}}{2} \|W-W_t\|^2_2) \\ 
\emph{s.t.} ~&H_t=Q_tU_t+\varepsilon_{1_t},\\
& Y_t=W_t(Q_tU_t)+\varepsilon_{2_t},\\
&\|\omega_{i}^T\|_2^2=1; \forall i=1,\dots,r,\\
&\|\omega_{t_i}^T\|_2^2=1; \forall i=1,\dots,r, \forall t=1,\dots,N,
\end{split}
\end{equation}
where $t$ represents the $t$th independent cluster, $\Omega_t$, $U_t$, $Q_t$ and $W_t$ are respectively the local analysis dictionary, sparse representation, structuring transformation and classifier of the $t$th cluster, and $\Omega$, $Q$, $W$ are respectively the global analysis dictionary, structuring transformation and classifier. The global variables will be applied to the same efficient classification scheme as the one of SADL.

\section{Algorithmic Solution} 
\label{sec:solve}
\subsection{SADL Algorithm}
Due to the non-convexity of the objective function in Eq. (\ref{equ:ssadl}), an augmented Lagrange formulation with dual variables $Z^{(1)}$, $Z^{(2)}$ and $\mu$ is adopted to seek an optimal solution. 
The augmented Lagrangian is then written as,
\begin{equation}\label{equ:Lssadl}
\begin{split}
&L(\Omega, U, Q, W, Z^{(1)}, Z^{(2)}, \mu)=\frac{1}{2}\|U - \Omega X\|_F^2 +\lambda_1 \|U\|_1  \\
&+  \langle Z^{(1)}, H-QU-\varepsilon_1\rangle+ \langle Z^{(2)},Y-W(QU)-\varepsilon_2 \rangle\\
&+\frac{\mu}{2}\|H-QU-\varepsilon_1\|_2^2+\frac{\mu}{2}\|Y-W(QU)-\varepsilon_2\|_2^2 \\
&+\frac{\rho_1}{2}\|\varepsilon_1\|^2_2 +\frac{\rho_2}{2}\|\varepsilon_2\|^2_2+\frac{\delta_1}{2}\|Q\|^2_2 +\frac{\delta_2}{2}\|W\|^2_2+\frac{\lambda_2}{2}\|\Omega\|^2_2,
\end{split}
\end{equation}
where $\lambda_1>0$ is a tuning parameter. 
To iteratively seek the optimal solution in Eq. (\ref{equ:Lssadl}), the analysis dictionary $\Omega$ and two linear transformations $Q$ and $W$ are first randomly initialized. The sparse representation $U$ is initialized as $U=\textbf{0}$, the zero matrix. In the following equations, Eq. (\ref{equ:updateU}) - Eq. (\ref{equ:updateY2}), the auxiliary variables $\eta_{U}$, $\eta_{Q}$, and $\eta_{W}$ are introduced to guarantee the convergence of the algorithm. The variables with superscripts which do not include parenthesis are the temporal variables of intermediate step in the calculation. Different variables are alternately updated while fixing the others, resulting in the following steps:\par

\noindent\textbf{(1)} Fix $\Omega$, $Q,~W$, and $\varepsilon_{1},~\varepsilon_2$, and update $U$: 
\begin{equation}\label{equ:updateU}
U_{k+1}=\tau_{\frac{\lambda_1}{\mu\eta_U}}(U_k-\frac{U_k^{1}+ U_k^{2}+ U_k^{3}}{\mu \eta_U}),
\end{equation}
where $\tau(\cdot)$ is the element-wise soft thresholding operator, and $U_k^{1}$, $U_k^{2}$, and $U_k^{3}$ are as follows:
\begin{equation}
U_k^{1}=-(\Omega_kX-U_k),
\end{equation}
\begin{equation}
U_k^{2}=-Q_k^T(Z_k^{(1)}+\mu(H-Q_kU_k-\varepsilon_{1_k})),
\end{equation}
\begin{equation}
U_k^{3}=-Q_k^TW_k^T(Z^{(2)}_k+\mu(Y-W_kQ_kU_k-\varepsilon_{2_k})).
\end{equation}

\noindent\textbf{(2)} Fix $\Omega$, $U, ~W$, and $\varepsilon_{1},~\varepsilon_2$, and update $Q$:
\begin{equation}\label{equ:updateQ}
Q_{k+1}=Q_k-\frac{ Q_k^{1}+  Q_k^{2} }{\mu\eta_Q},
\end{equation}
\begin{equation}
Q_k^{1}=-(Z_k^{(1)}+\mu(H-Q_kU_{k+1}-\varepsilon_{1_k}))U_{k+1}^T+\delta_1Q_k,
\end{equation}
\begin{equation}
Q_k^{2}=-W_k^T(Z_k^{(2)}+\mu(Y-W_kQ_kU_{k+1}-\varepsilon_{2_k}))U_{k+1}^T.
\end{equation}
\noindent\textbf{(3)} Fix $\Omega$, $U, ~Q$, and $\varepsilon_{1},~\varepsilon_2$, and update $W$:
\begin{equation}\label{equ:updateW}
W_{k+1}=W_k-\frac{W_k^{1} }{\mu\eta_W}
\end{equation}
\begin{equation}
\resizebox{1\columnwidth}{!}{$W_k^{1}=-(Z_k^{(2)}+\mu(Y-W_kQ_{k+1}U_{k+1}-\varepsilon_{2_k}))U_{k+1}^TQ_{k+1}^T+\delta_2W_k.$}
\end{equation}

\noindent\textbf{(4)} Fix $U$, $Q,~W$, and $\varepsilon_{1},~\varepsilon_2$, and update $\Omega$:
\begin{equation}\label{equ:adl-d}
\Omega_{k+1}^*=\arg\min_\Omega \frac{1}{2}\|U_{k+1}-\Omega X\|_F^2.
\end{equation}
The analytical solution of Eq. (\ref{equ:adl-d}) can be regularized as
\begin{equation}\label{equ:updateD}
\Omega_{k+1}=U_{k+1}X^T(XX^T+\lambda_2 I)^{-1},
\end{equation}
where $\lambda_2$ is also a tuning parameter. It will be chosen by a usual way.

\noindent\textbf{(5)} Fix $U$, $\Omega$, $Q, ~W$, and $~\varepsilon_2$, and update $\varepsilon_{1}$:
\begin{equation}\label{equ:adl-epsilon1}
\varepsilon_{1_{k+1}}=\frac{1}{\rho_1-1}(Z_k^{(1)}+\mu(H-Q_{k+1}U_{K+1})) .
\end{equation}

\noindent\textbf{(6)} Fix $U$, $\Omega$, $Q, ~W$, and $~\varepsilon_1$, and update $\varepsilon_{2}$:
\begin{equation}\label{equ:adl-epsilon2}
\varepsilon_{2_{k+1}}=\frac{1}{\rho_2-1}(Z_k^{(2)}+\mu(Y-W_{k+1}Q_{k+1}U_{K+1})) .
\end{equation}

\noindent The dual variables $Z^{(1)}$, $Z^{(2)}$ are updated as
\begin{equation}\label{equ:updateY1}
Z_{k+1}^{(1)}=Z_{k}^{(1)}+\mu(H-Q_{k+1}U_{k+1}),
\end{equation}
\begin{equation}\label{equ:updateY2}
Z_{k+1}^{(2)}=Z_{k}^{(2)}+\mu(Y-W_{k+1}Q_{k+1}U_{k+1}).
\end{equation}

In contrast to previous ADL techniques, which train a dictionary by iterating a single row of the dictionary, \emph{i.e.,} one atom, to avoid a trivial solution, we proceed to update a set of rows in a single step at each iteration. A fast convergence rate of the algorithm is also guaranteed by linearized ADM \cite{lin2011linearized} and with a  closed form solution for the dictionary $\Omega$ given in Eq. (\ref{equ:updateD}). The proposed SADL algorithm \footnote{The codes are released at https://github.com/wtang0512/Demo-of-SADL} is summarized in Algorithm 1.\par

\renewcommand{\algorithmicrequire}{\textbf{Input:}}
\renewcommand{\algorithmicensure}{\textbf{Output:}}
\begin{algorithm} \label{alg:learning0}  
	\caption{Structured Analysis Dictionary Learning}  
	\begin{algorithmic}[1] 
		\Require 
		Training data $X=[x_1,\dots,x_n]$, diagonal block matrix $H$, class labels $Y$, penalty coefficients $\rho_1,\rho_2,\delta_1,\delta_2$, parameters $\lambda_1,\lambda_2$ and maximum iteration $p$;
		\Ensure 
		Analysis dictionary $\Omega$, sparse representation $U$, and linear transformations $Q$ and $W$; 
		\State Initialize $\Omega$, $Q$, and $W$ as random matrices, and initialize $U$ as a zero matrix;
		\While {not converged \textbf{and} $k < p$}
		\State $k=k+1$;
		\State Update $U_{k}$ by (\ref{equ:updateU});
		\State Update $Q_{k}$ by (\ref{equ:updateQ});
		\State Update $W_{k}$ by (\ref{equ:updateW});
		\State Update $\Omega_{k}$ by (\ref{equ:updateD});
		\State Update $\varepsilon_{1_k}$ by (\ref{equ:adl-epsilon1});
		\State Update $\varepsilon_{2_k}$ by (\ref{equ:adl-epsilon2});
		\State Update $Z^{(1)}_{k}$ by (\ref{equ:updateY1});
		\State Update $Z^{(2)}_{k}$ by (\ref{equ:updateY2});
		\EndWhile
	\end{algorithmic}  
\end{algorithm}  

\subsection{Distributed SADL Algorithm}
The distributed SADL is similarly expressed in the augmented Lagrangian function as
\begin{equation}\label{equ:sadlp}
\begin{split}
&L_d(\Omega_t, U_t, Q_t, W_t,\Omega,Q,W, Z^{(1)}, Z^{(2)}, \mu_k)=\\
&\sum_{t=1}^N (\frac{1}{2}\|U_t - \Omega_t X_t\|_F^2 +\lambda_1 \|U_t\|_1 +\frac{\delta_{1_t}}{2}\|Q_t\|^2_2 +\frac{\delta_{2_t}}{2}\|W_t\|^2_2  \\
&+\frac{\xi_{1_t}}{2}\|\Omega-\Omega_t\|^2_2 +\frac{\xi_{2_t}}{2} \|Q-Q_t\|^2_2+\frac{\xi_{3_t}}{2} \|W-W_t\|^2_2\\
& +\frac{\rho_{1_t}}{2}\|\varepsilon_{1_t}\|^2_2+\frac{\rho_{2_t}}{2}\|\varepsilon_{2_t}\|^2_2+\frac{\lambda_{2_t}}{2}\|\Omega_{t}\|^2_2\\
&+ \langle Z^{(1)}_t, H_t-Q_tU_t-\varepsilon_{1_t}\rangle+ \langle Z^{(2)}_t,Y_t-W_t(Q_tU_t)-\varepsilon_{2_t}\rangle\\
&+\frac{\mu_k}{2}\|H_t-Q_tU_t-\varepsilon_{1_t}\|_2^2+\frac{\mu_k}{2}\|Y_t-W_t(Q_tU_t)-\varepsilon_{2_t}\|_2^2).
\end{split}
\end{equation}
To minimize such an objective function, each variable is alternatively updated while fixing others. The distributed SADL algorithm is presented in Algorithm 2.

\begin{algorithm} \label{alg:distributedlearning}  
	\caption{Distributed SADL}  
	\begin{algorithmic}[2] 
		\Require 
		Training data $X=[x_1,\dots,x_n]$, diagonal block matrix $H$, class labels $L$, penalty coefficients $\delta_{1_t},\delta_{2_t},\xi_{1_t},\xi_{2_t},\xi_{3_t}$, parameters $\lambda_1,\lambda_2$ and maximum iteration $p$;
		\Ensure 
		Analysis dictionary $\Omega$, linear transformations $Q$ and $W$; 
		\State Initialize  $\Omega_t$, $Q_t$, $W_t$, $\Omega$, $Q$, and $W$ as random matrices, initialize $U_t$ as a zero matrix, and set $\{X_t\}$ as a randomly selected partition of $X$ with $\bigcup_{t=1}^N X_t=X$;
		\While {not converged \textbf{and} $k < p$}
		\State $k=k+1;$
		\For {$t=1:N$} \%Here for loop can be parallelized or distributed in different clusters.
		\State \resizebox{0.85\columnwidth}{!}{$U_{t}^{k+1}=\tau_{\frac{\lambda_1}{\mu\eta_U}} \left(U_t^k-\frac{\bigtriangledown_UL_d(\Omega_t^k,U_t^k,Q_t^k,W_t^k,\Omega^k,Q^k,W^k,Y^{(1)^k}_t,Y^{(2)^k}_t)}{\mu \eta_U}\right);$}
		\State \resizebox{0.85\columnwidth}{!}{$Q_t^{k+1}=Q_t^k-\frac{\bigtriangledown_QL_d(\Omega_t^k,U_t^{k+1},Q_t^k,W_t^k,\Omega^k,Q^k,W^k,Y^{(1)^k}_t,Y^{(2)^k}_t) }{\mu \eta_Q};$}
		\State \resizebox{0.85\columnwidth}{!}{$W_t^{k+1}=W_t^k-\frac{\bigtriangledown_WL_d(\Omega_t^k,U_t^{k+1},Q_t^{k+1},W_t^k,\Omega^k,Q^k,W^k,Y^{(1)^k}_t,Y^{(2)^k}_t) }{\mu\eta_W};$}
		\State \resizebox{0.85\columnwidth}{!}{$\Omega_t^{k+1}=(U_t^{k+1}X_t^T+\xi_{1_t} \Omega^k)(X_tX_t^T+ (\xi_{1_t}+\lambda_{2_t}) I)^{-1};$}
		\State Normalize $\Omega_t^{k+1}$ by $\omega_{t_i}^T=\frac{\omega_{t_i}^T}{\|\omega_{t_i}^T\|_2}, \forall i;$
		\State $Y_{k+1}^{(1)}=Y_{k}^{(1)}+\mu(H-Q_{k+1}U_{k+1});$
		\State $Y_{k+1}^{(2)}=Y_{k}^{(2)}+\mu(L-W_{k+1}Q_{k+1}U_{k+1});$
		\State $\mu_{k+1}=\min\{\rho\mu,\mu_{max}\};$ 
		\State $\xi_{1_{k+1}}=\min\{\rho\xi_{1_{k}},\xi_{1_{max}}\};$ 
		\State $\xi_{2_{k+1}}=\min\{\rho\xi_{2_{k}},\xi_{2_{max}}\};$ 
		\State $\xi_{3_{k+1}}=\min\{\rho\xi_{3_{k}},\xi_{3_{max}}\};$ 
		\EndFor
		\State $\Omega^{k+1}=\frac{1}{N}\sum_t \Omega_t^{k+1};$
		\State Normalize $\Omega^{k+1}$ by $\omega_{i}^T=\frac{\omega_{i}^T}{\|\omega_{i}^T\|_2}, \forall i;$
		\State $Q^{k+1}=\frac{1}{N}\sum_t Q_t^{k+1};$
		\State $W^{k+1}=\frac{1}{N}\sum_t W_t^{k+1};$
		\EndWhile
	\end{algorithmic}  
\end{algorithm}

\section{Classification Procedure} \label{sec:classifier}
\begin{figure}[htb]
	\centering
	\includegraphics[width=0.45\textwidth]{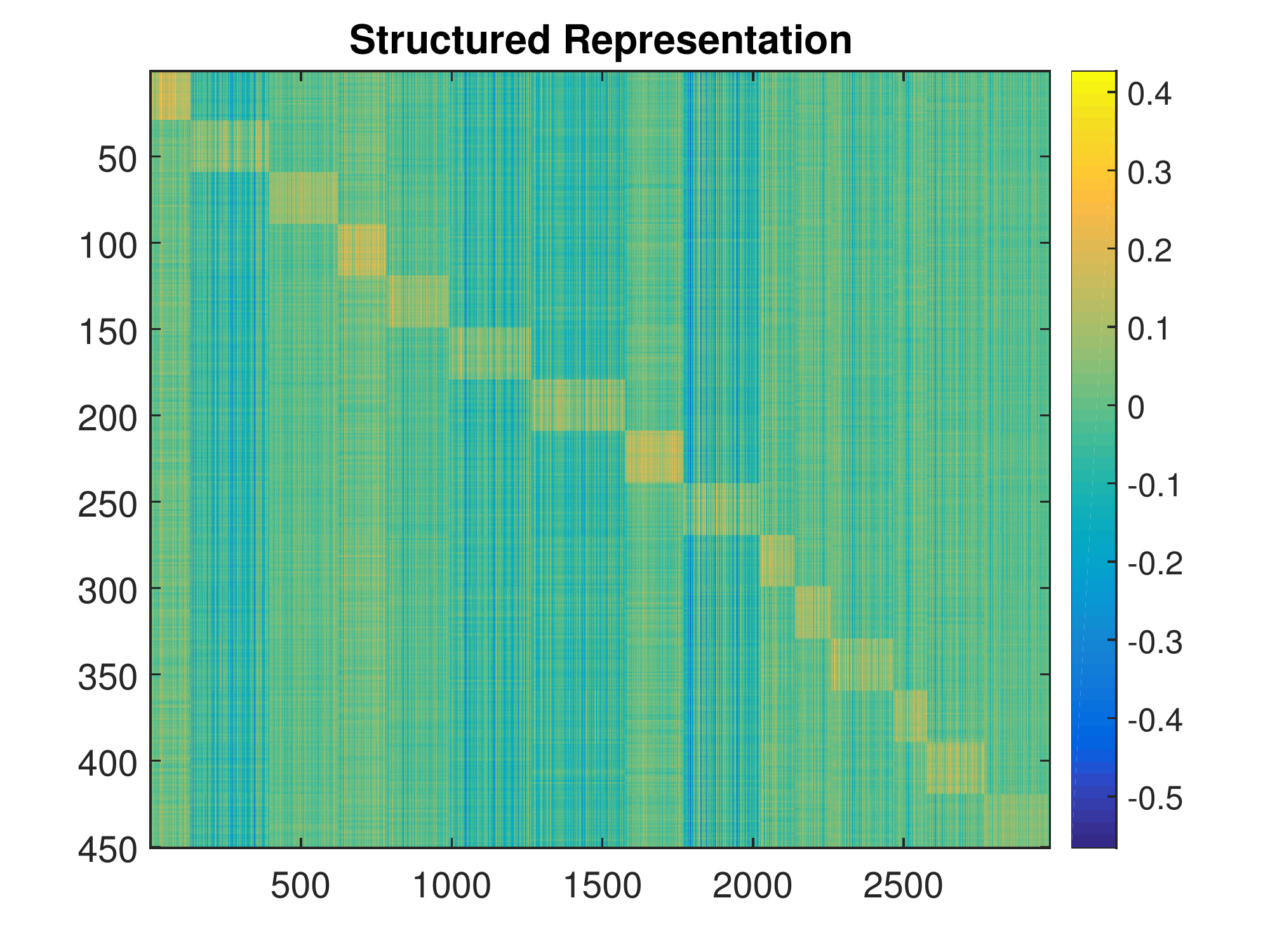}
	\caption{$Q(\Omega x)$ on Scene 15 Dataset}
	\label{fig:diag}
\end{figure}

Both SADL and Distributed SADL have the same classification procedure because the global analysis dictionary $\Omega$, transforming matrix $Q$ and classifier $W$ are obtained from the algorithms. With the analysis dictionary $\Omega$ in hand, an observed image $x$ can be quickly sparsely encoded as $\Omega x$. This is in stark contrast to SDL for which a sparse representation is obtained by solving a non-smooth optimization as: $\arg\min_\alpha \|x-D\alpha\|_2^2+\|\alpha\|_1,$ and highlights the remarkable improvement ADL provides. Our proposed SADL, which naturally enjoys the same encoding properties as ADL, efficiently yields a structured sparse representation $Q(\Omega x)$ of the signal $x$ as well. Figure \ref{fig:diag} shows an example of the structured representations obtained from Scene 15 dataset.
\begin{figure}[htb]
	\centering
	\includegraphics[width=0.45\textwidth]{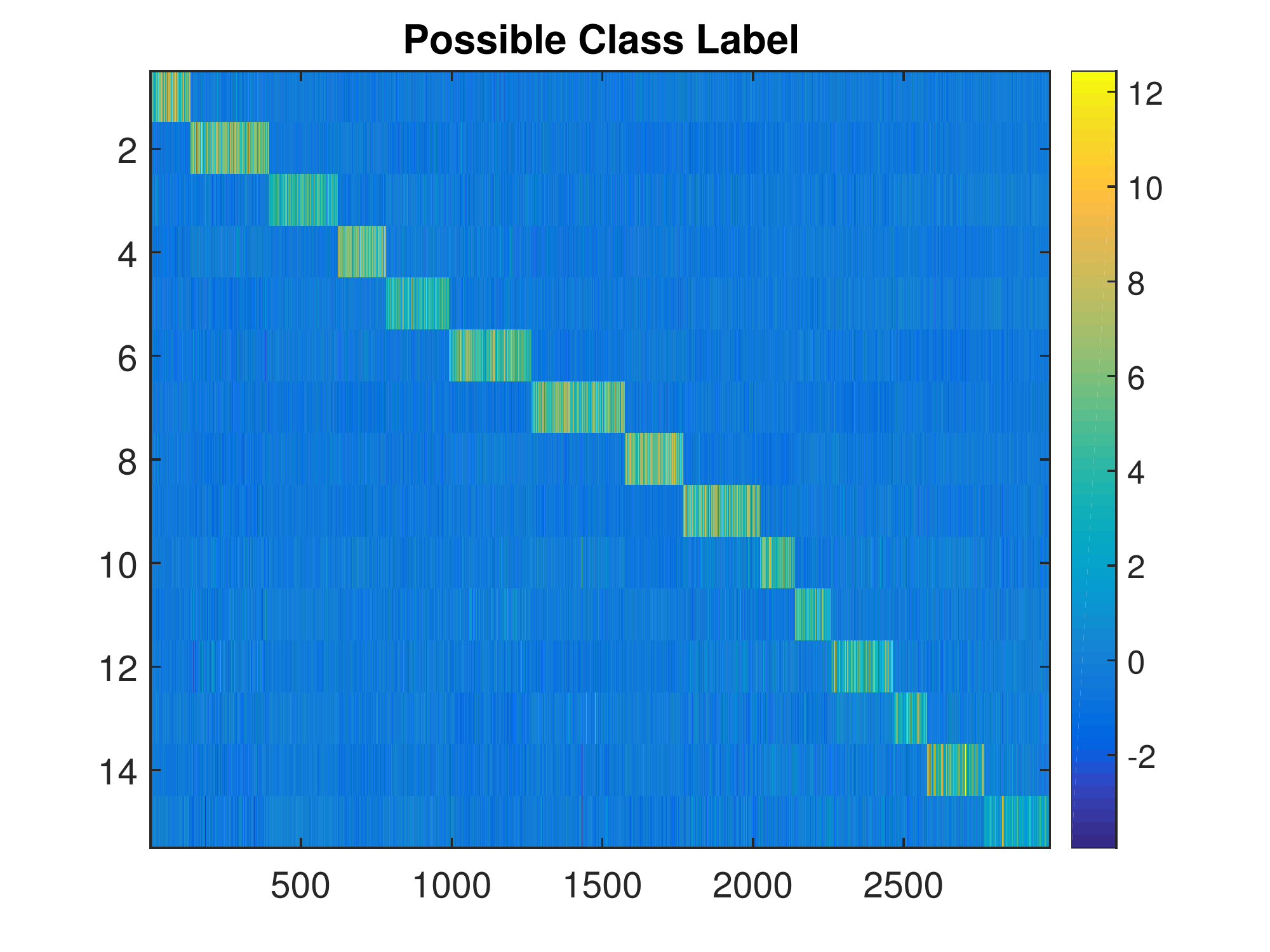}
	\caption{$WQ(\Omega x)$ on Scene 15 Dataset}
	\label{fig:classlabel}
\end{figure}
As shown, the result reflects the desired block diagonal structure. The ultimate desired classification goal of $x$ is accomplished by $W(Q \Omega x)$. Figure \ref{fig:classlabel} depicts $W(Q\Omega x)$ for the example in Figure \ref{fig:diag} where the horizontal axis is image index, and the vertical axis reflects the class labels, which are computed according to, 
\begin{equation}
y=\max_j (WQ \Omega x)_j,
\end{equation}
shown as the brightest ones in Figure \ref{fig:classlabel}.

\section{Convergence} \label{sec:Convergeandcomplexity}
Since we have used linearized ADM method to solve our nonconvex objective function, $\eta_U, \eta_Q, \eta_W$ are introduced as the auxiliary variables. We additionally have the following
\begin{theorem}\label{theorem:main}
	Suppose that $\mu\geq\sqrt 2\{\rho_1,\rho_2\}$. There exist positive values $\eta^0_U,\eta^0_Q,\eta^0_W, R$ only depending on the initialization such that for $\eta_U>\eta^0_U, \eta_Q>\eta^0_Q, \eta_W>\eta^0_W$ the sequence $\{\Theta_k=(\Omega_k,U_k,Q_k,W_k,\varepsilon_k^{(1)},\varepsilon_k^{(2)}, Z_k^{(1)}, Z_k^{(2)})\}_{k=1}^\infty$ converges to the following set of bounded feasible stationary points of the Lagrangian \footnote{The norm $\|\Theta\|$ is any norm that is continuous with respect to the two norm of the components, for example their some of two norms. Also, the function $\|U\|_1$ is treated as a (convex) function of $\Theta$, which is constant with respect to other components than $U$.}:
	\[S=\{\Theta=(\Omega,U,Q,W,\varepsilon^{(1)},\varepsilon^{(2)}, Z^{(1)}, Z^{(2)})\mid\]
	\[ \|\Theta\|<R, -\nabla L_s\in\lambda\partial\|U\|_1,H=QU+\varepsilon^{(1)}, Y=QUW+\varepsilon^{(2)}\}\]
	where $L_s$ is the smooth part of $L$, \emph{i.e.,}
	\[
	L=L_s+\lambda_1\|U\|_1.
	\]
\end{theorem}
According to Theorem 1, if we initialize $\eta_U, \eta_Q, \eta_W$ large enough, Algorithm 1 not only converges, but also generates the variable sequences with a final convergence to the stationary points. The proof of Theorem 1 can be found in Appendix A.

\section{Experiments and Results}\label{sec:experiments}
We now evaluate our proposed SADL method on five popular visual classification datasets that have been widely used in previous works and with known performance benchmarks. They include Extended YaleB face dataset \cite{yaleB}, AR face dataset \cite{AR}, Caltech 101 object categorization dataset \cite{caltech101}, Caltech 256 objective dataset \cite{griffin2007caltech}, and  Scene 15 scene image dataset \cite{scene15}.\par

In our experiments, we provide a comparative evaluation of six state-of-the-art techniques and our proposed technique, including a classification accuracy as well as training and testing times. All our experiments and competing algorithms are implemented in Matlab 2015b on the server with 2.30GHz Intel(R) Xeon(R) CPU. For a fair comparison, we measure the performance of each algorithm by repeating the experiment over 10 realizations. 
The testing time is defined as the average processing time to classify a single image. In our tables, the accuracy in parentheses with the associated citation is that was reported in the original paper. The difference in the accuracy of our approach and of the original one might be caused by different segmentations of the training and testing samples.\par

\subsection{Parameter Settings}
In our proposed SADL method, $\lambda_1,\lambda_2$ and maximum iteration $p$ are tuning parameters. The parameter $\lambda_1$ controls the contribution of the sparsity, and the parameter $\lambda_2$ controls the learned analysis dictionary, while $p$ is the maximum iteration number. We replace $\varepsilon_1$ and $\varepsilon_2$ by their expressions, and insert them in the optimization formula.
We choose for all the experiments $\lambda_1,\lambda_2$, $p$ and dictionary size by 10-fold cross validation on each dataset. In addition, we also optimally tuned the parameters of all competing methods to ensure their best performance.

\subsection{State-of-the-art Methods}
We compare our proposed SADL and Distributed SADL (DSADL) with the following competing techniques: The first one is a baseline, which uses the ADL method to learn a sparse representation and subsequently trains a Support Vector Machine (SVM) to classify images based on such sparse representations (ADL+SVM) \cite{shekhar2014analysis}. A penalty term is included to avoid similar atoms and minimize false positives. The second one is the classical Sparse Representation based Classification (SRC) \cite{src}. For this method, we do not need to train a dictionary. Instead, we use the training images as the atoms in the dictionary. In the testing phase, we obtain the sparse coefficients based on such a dictionary. The third technique that we consider in this work is a state-of-the-art dictionary learning method, called Label Consistent K-SVD (LC-KSVD) \cite{lcksvd}, which forces each category labels to be consistent with classification. We select the LC-KSVD2 in \cite{lcksvd} for comparison, because it has a better classification performance. The fourth method is Discriminative Analysis Dictionary Learning (DADL) \cite{dadl}, which incorporates a topological structure and distinct class representations to the ADL framework in order to make each class discriminative. Then a 1-nearest-neighbor classifier is used to assign the label. The fifth technique, Class-aware Analysis Dictionary Learning (CADL) \cite{cadl}, is to learn the class-specific analysis dictionaries and jointly learn a universal SVM based on the concatenated class-specific coefficients of each class. Finally, we compare our method with the Synthesis K-SVD based Analysis Dictionary Learning (SK-SVDADL) \cite{sksvdadl}, which is to jointly learn ADL and a linear classifier and is solved by the K-SVD method.

\subsection{Extended YaleB}
\begin{figure}[htb]
	\centering
	\includegraphics[width=0.45\textwidth]{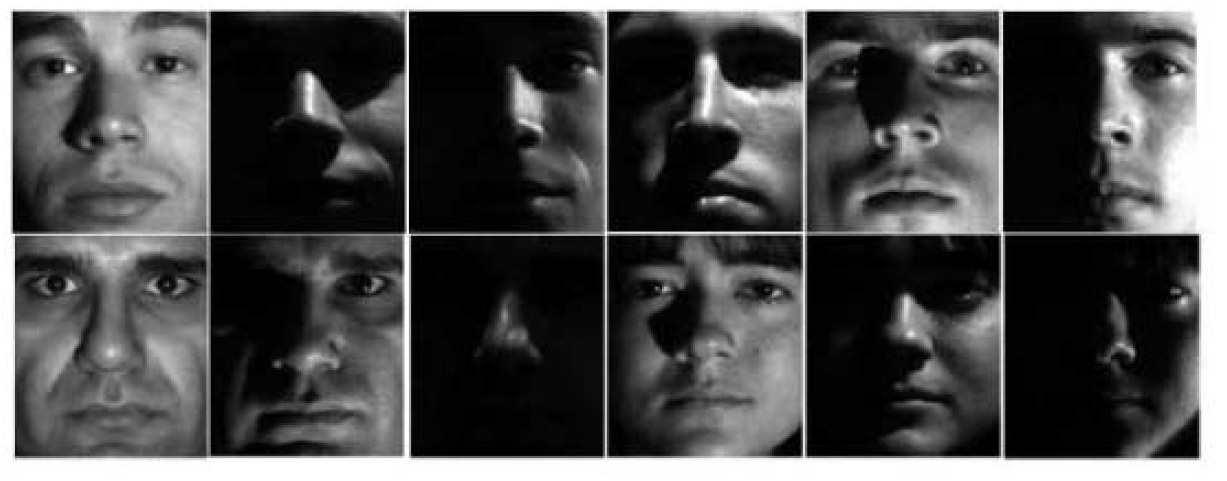}
	\caption{Extended YaleB Dataset Examples}
	\label{fig:YaleB}
\end{figure}
The Extended YaleB face dataset contains in total 2414 frontal face images of 38 persons under various illumination and expression conditions, as illustrated in Figure \ref{fig:YaleB}. Due to such illumination and expression variation, YaleB is intended to test the robustness to the intra-class variation. Each person has about 64 images, each cropped to $168 \times 192$ pixels. We project each face image onto a n-dimensional random face feature vector. The projection is performed by a randomly generated matrix with a zero mean normal distribution whose rows are $l_2$ normalized. This procedure is similar to the one in \cite{lcksvd}. In our experiment, $n$ is 504, \emph{i.e.,} each Extended YaleB face image is reduced to a $504$-dimensional feature vector. Then, we randomly choose half of the images for training, and the rest for testing. The dictionary size is set to 1216 atoms, $\lambda_1=0.001,\lambda_2=0.005$ and $p=466$.

\begin{table}[htb]
	\centering
	\caption{Classification Results on Extended YaleB Dataset}
	\begin{tabular}{llll}
		\hline
		Methods & Classification & Training & Testing \\
		& Accuracy(\%) & Time(s) & Time(s)\\
		\hline
		ADL+SVM\cite{shekhar2014analysis} & 82.91\% & 91.78 & 1.13$\times 10^{-3}$\\
		SRC\cite{src} & 96.51\% & No Need & 3.66$\times 10^{-1}$\\
		LC-KSVD\cite{lcksvd} & 83.31\% ({96.7\%}\cite{lcksvd}) & 123.07 & 1.60$\times 10^{-2}$\\
		DADL\cite{dadl} & \textbf{97.35}\% (97.7\%\cite{dadl}) & \textbf{10.05} & 4.55$\times 10^{-5}$\\
		CADL\cite{cadl} & {97.05}\% & 130.83 & 9.72$\times 10^{-6}$\\
		SK-SVDADL\cite{sksvdadl} & 96.14\% (96.9\%\cite{sksvdadl}) &113.78& 1.34$\times 10^{-4}$\\
		SADL & 96.35\% &{39.23} & \textbf{7.61$\times 10^{-6}$}\\
		\hline
	\end{tabular}%
	\label{tab:yaleb}%
\end{table}%

{The classification results, training and testing times are summarized in Table \ref{tab:yaleb}.
Although the accuracy of the SADL method is slightly lower than SRC, DADL and CADL, it is still comparable and higher than SK-SVADL, LC-KSVD and ADL+SVM. SADL is substantially more efficient than the others in terms of numerical complexity. }\par

\begin{figure}[htb]
	\centering
	\includegraphics[width=0.45\textwidth]{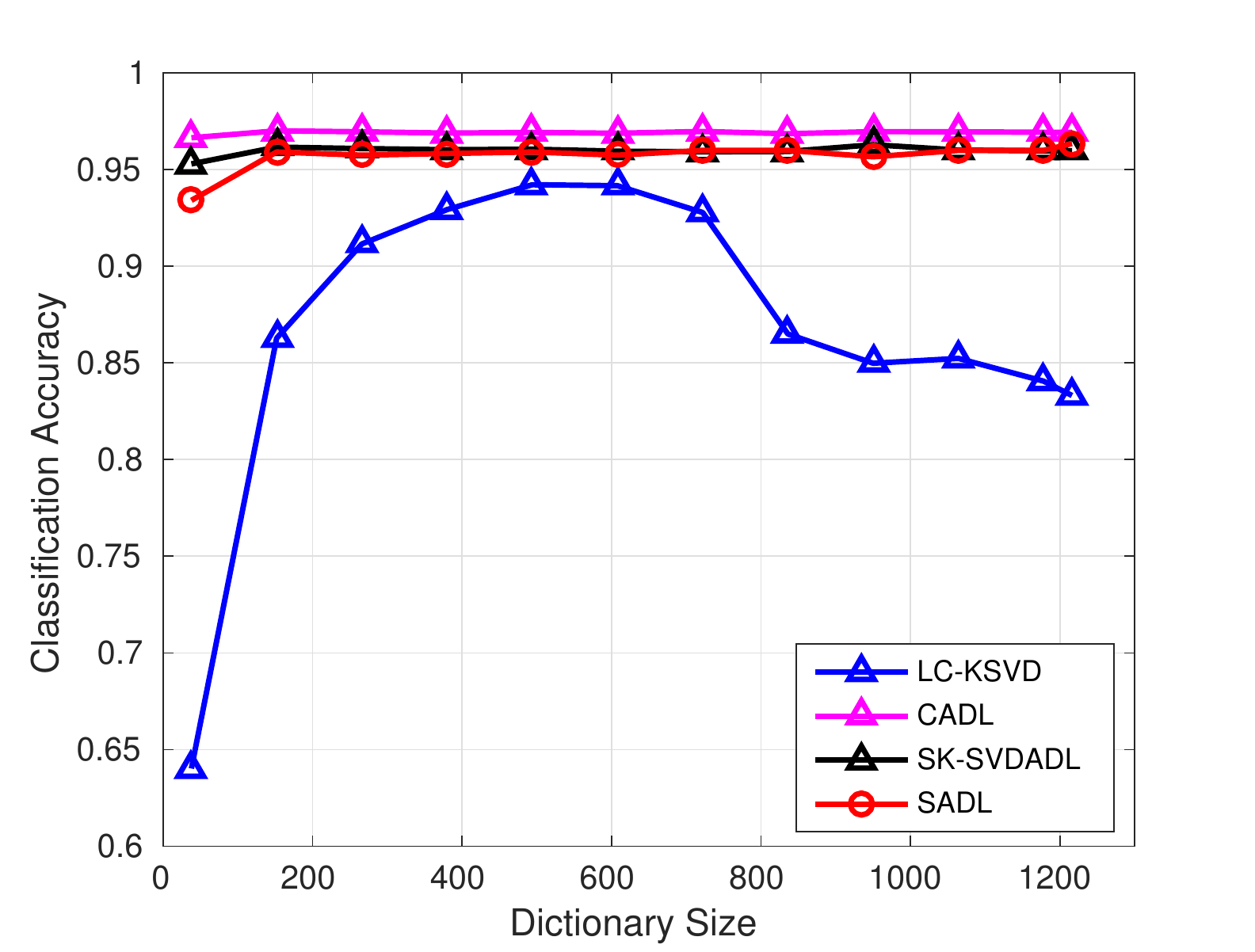}
	\caption{Classification Accuracy versus Dictionary Size}
	\label{fig:dictsize}
\end{figure}

\begin{figure}[htb]
	\centering
	\includegraphics[width=0.45\textwidth]{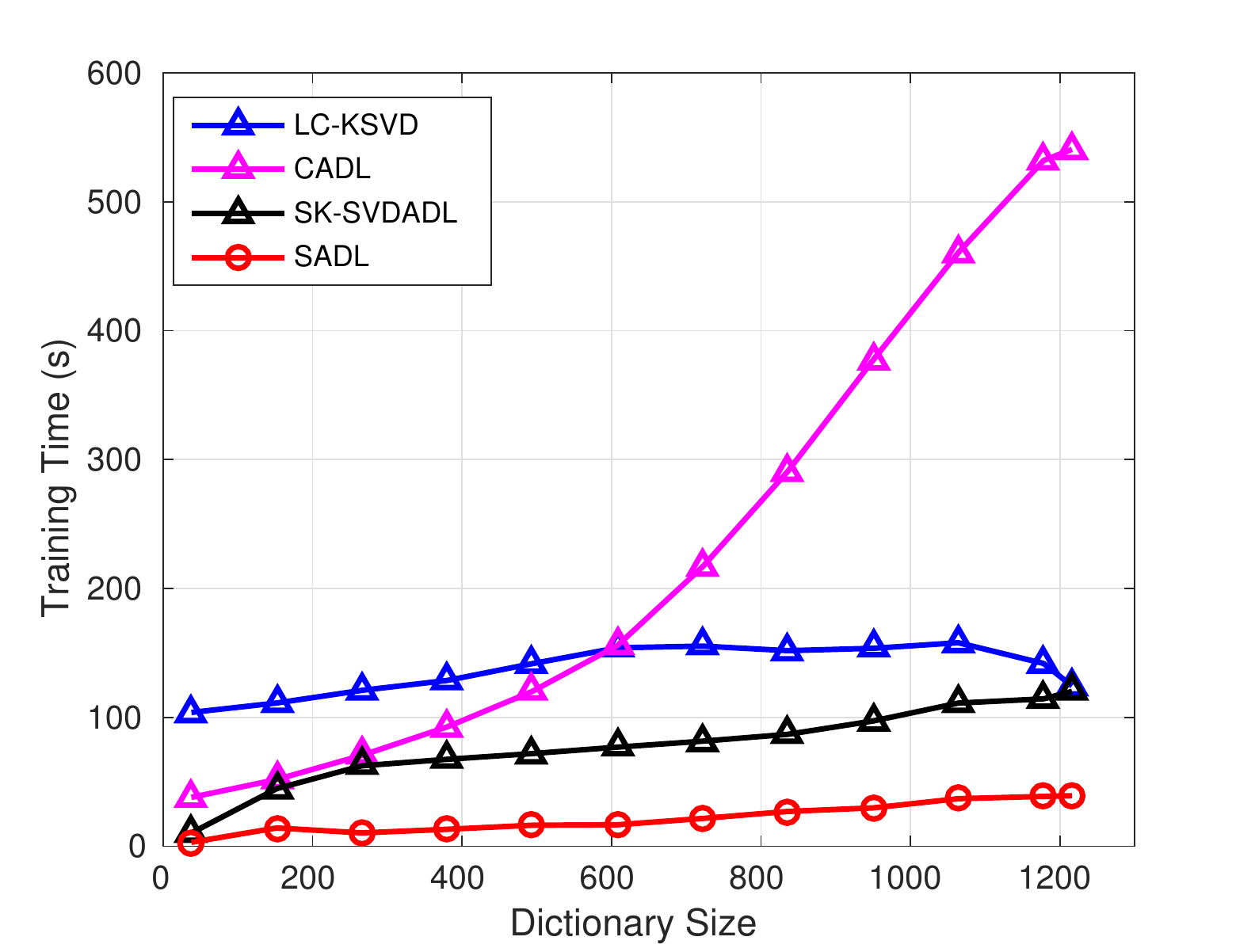}
	\caption{Training Time versus Dictionary Size}
	\label{fig:traintime}
\end{figure}

For a more thorough evaluation, we compare SADL with LC-KSVD, CADL and SK-SVDADL for different dictionary sizes, and display the classification accuracy and training times in Figure \ref{fig:dictsize} and \ref{fig:traintime}, which are based on the average of ten realizations. We ran our experiments for dictionary sizes by the size of 38, 152, 266, 380, 494, 608, 722, 836, 950, 1064, 1178 and 1216 (all training size). SADL, SK-SVDADL and CADL, the ADL methods, exhibit a more stable accuracy performance than that of LC-KSVD of the SDL methods. In particular, the accuracy of LC-KSVD significantly decreases, when the dictionary size approaches the training sample size. The significant decrease in accuracy may be caused by the trivial solution of dictionary $D$ in SDL. In addition, our method apparently has a much higher classification accuracy than LC-KSVD and a very similar accuracy as SK-SVDADL, when the dictionary size is small. As the dictionary size increases, SADL achieves a better accuracy than SK-SVDADL and approaches the accuracy of CADL. Although the accuracy of SADL is barely lower than CADL, the SADL method is also much faster than the LC-KSVD, SK-SVDADL and CADL in the training phase, especially when the dictionary size becomes larger. \par

\subsection{AR Face}
\begin{figure}[htb]
	\centering
	\includegraphics[width=0.45\textwidth]{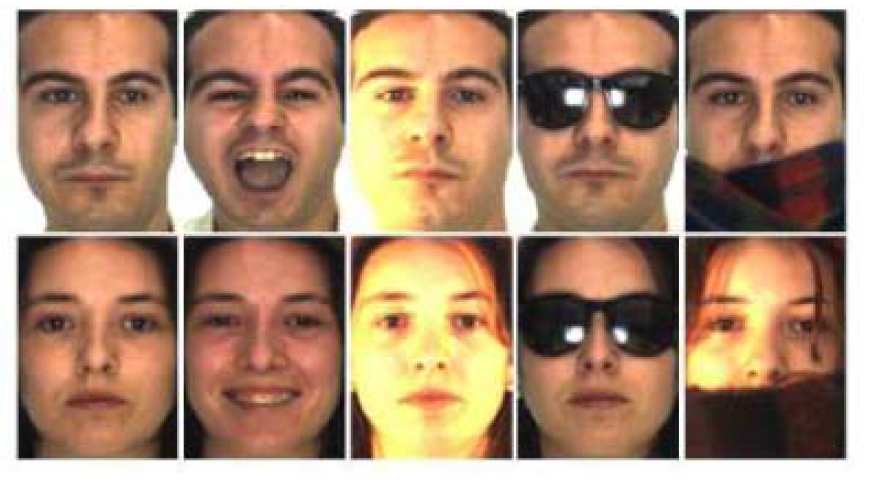}
	\caption{AR Dataset Examples}
	\label{fig:ar}
\end{figure}

The AR Face dateset has 2600 color images of 50 females and 50 males with more facial variations than the Extended YaleB database, such as different illumination conditions, expressions and facial disguises. AR is also used to test the robustness to large intra-class variation. Each person has about 26 images of size $165 \times 120$. Figure \ref{fig:ar} shows some sample images of faces with sunglasses or scarves. The features of the AR face image are extracted in the same way as those of the Extended YaleB face image are, but we project it to a $540$ dimensional feature vector similarly to the setting in \cite{lcksvd}. 20 images of each person are randomly selected as a training set and the other 6 images for testing. The dictionary size of the AR dataset is set to 2000 atoms, $\lambda_1=0.001$, $\lambda_2=0.005$ and $p=204$.

\begin{table}[htb]
	\centering
	\caption{Classification Results on AR Dataset}
	\begin{tabular}{llll}
		\hline
		Methods & Classification & Training & Testing \\
		& Accuracy(\%) & Time(s) & Time(s)\\
		\hline
		ADL+SVM\cite{shekhar2014analysis} & 90.40\% & 218.54 & 9.10$\times 10^{-3}$\\
		SRC\cite{src} & 97.10\% & No Need & 7.41$\times 10^{-1}$\\
		LC-KSVD\cite{lcksvd} & 87.78\% (97.8\%\cite{lcksvd}) & 169.35 & 2.00$\times 10^{-2}$\\
		DADL\cite{dadl} & 98.32\% (98.7\%\cite{dadl})& 47.76 & 2.68$\times 10^{-4}$\\
		CADL\cite{cadl} & \textbf{98.52}\% (98.8\%\cite{cadl})  & 313.37 & 1.34$\times 10^{-5}$\\
		SK-SVDADL\cite{sksvdadl} & 97.38\% (97.7\%\cite{sksvdadl}) &113.78& 1.34$\times 10^{-4}$\\
		SADL & {97.17\%} & \textbf{32.60} & \textbf{1.33$\times 10^{-5}$}\\
		\hline
	\end{tabular}%
	\label{tab:ar}%
\end{table}%
{The classification results as well as the training and testing times are summarized in Table \ref{tab:ar}. Comparing with other methods, our proposed SADL achieves a comparable result with the fastest training and testing time. The classification accuracy is lower than DADL, CADL and SK-SVDADL, and higher than SRC, LC-KSVD. However, our method is about 1000 times faster than SRC and LC-KSVD for the testing phase, 10 times faster than DADL and SK-SVDADL. Although SADL is only slightly faster than CADL, its training time is one-tenth of the one of CADL.}

\subsection{Caltech 101}
\begin{figure}[htb]
	\centering
	\includegraphics[width=0.45\textwidth]{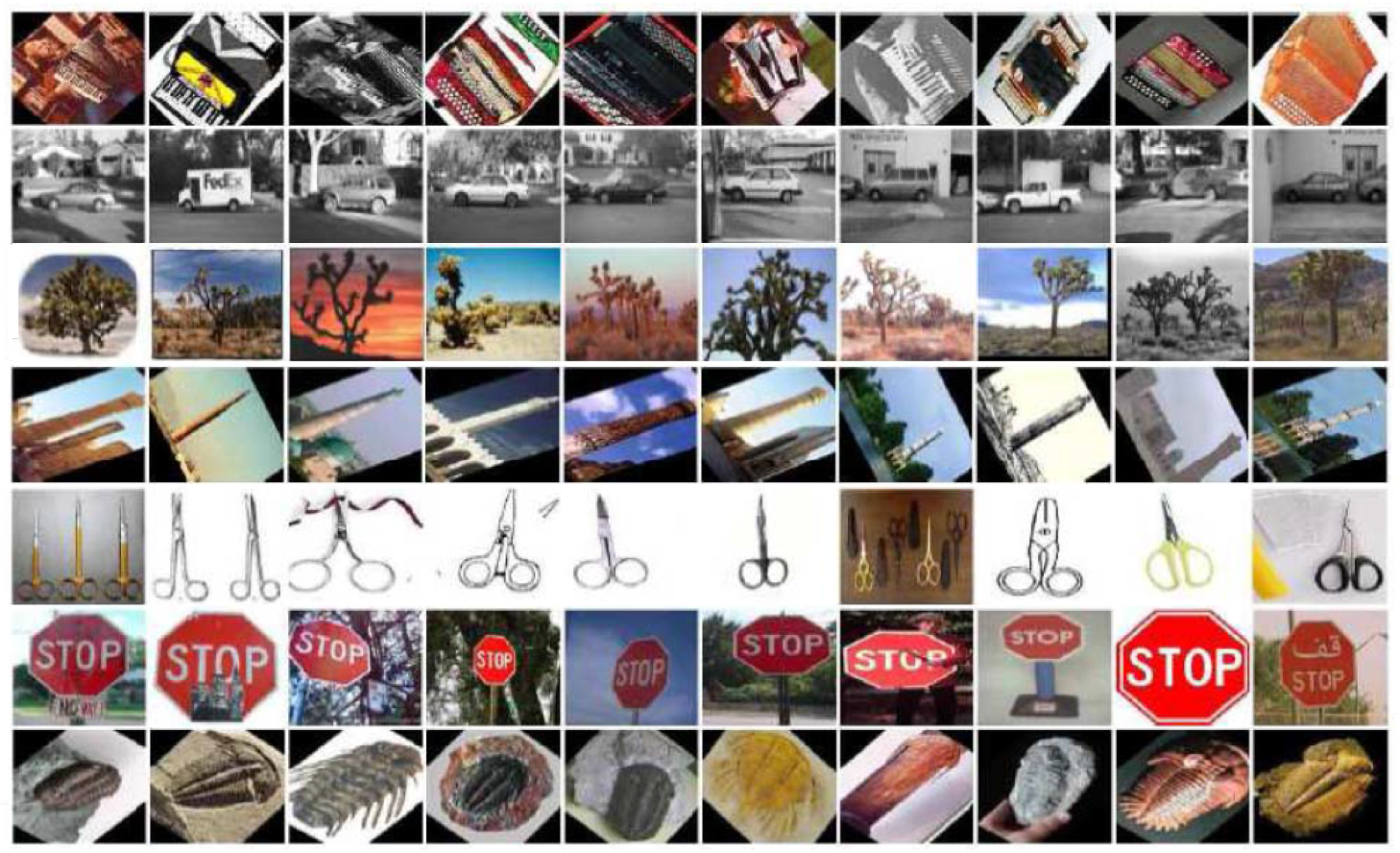}
	\caption{Caltech 101 Dataset Examples}
	\label{fig:caltech101}
\end{figure}

The Caltech 101 dataset has 101 different categories of different objects and one non-object category. Most categories have around 50 images. Figure \ref{fig:caltech101} gives some examples from the Caltech 101 dataset. Since this dataset is left-right aligned and rotated, Caltech 101 contians many different intra-class scaling variations, color pattern diversity and inter-class common features. We extract dense Scale-invariant Feature Transform (SIFT) descriptors for each image from $16 \times 16$ patches and with a $6$ pixels step. 
Then, we apply a spatial pyramid method \cite{scene15} to the dense SIFT features with three segmentation sizes $1 \times 1$, $2\times 2$, and $4\times 4$ to capture the objects' features at different scales. At the same time, a $1024$ size codebook is trained by $k$-means clustering for spatial pyramid features. 
Spatial pyramid features of each subregion are then concatenated together as a vector to represent one image. Due to the sparse nature of the spatial pyramid features, we use PCA to reduce each feature to $3000$ dimensions. In our experiment, 30 images per class are randomly chosen as training data, and other images are used as testing data. All the steps and settings follow \cite{lcksvd}. The dictionary size is set to 3060, $\lambda_1=0.001,\lambda_2=1.5$ and $p=827$. 

\begin{table}[htb]
	\centering
	\caption{Classification Results on Caltech 101 Dataset}
	\begin{tabular}{llll}
		\hline
		Methods & Classification & Training & Testing \\
		& Accuracy(\%) & Time(s) & Time(s)\\
		\hline
		ADL+SVM\cite{shekhar2014analysis} & 66.75\% &{1943.47}  & 1.33$\times 10^{-2}$\\
		SRC\cite{src} & 70.70\% & No Need & 4.34$\times 10^{-1}$\\
		LC-KSVD\cite{lcksvd} & 73.67\% (73.6\%\cite{lcksvd}) & 2144.90 & 2.49$\times 10^{-3}$\\
		DADL\cite{dadl} & 71.77\% (74.6\%\cite{dadl}) & 233.49 & 7.90$\times 10^{-4}$\\
		CADL\cite{cadl} & \textbf{76.83}\% (75\%\cite{cadl})  & 9896.46 & 4.86$\times 10^{-5}$\\
		SK-SVDADL\cite{sksvdadl} & 73.39\% (74.4\%\cite{sksvdadl}) &\textbf{182.71}& 2.49$\times 10^{-4}$\\
		SADL & {74.45\%} & {847.50} & \textbf{4.76$\times 10^{-5}$}\\
		\hline
		DADL & 73.49\% & - & \textbf{8.10$\times 10^{-6}$}\\
		\hline
	\end{tabular}%
	\label{tab:caltech101}%
\end{table}%

{The classification results, training and testing times are summarized in Table \ref{tab:caltech101}. Our proposed SADL achieves the second highest accuracy, while only costing one-tenth of the training time of CADL obtaining the maximum accuracy. SADL has again the shortest encoding time, which is around 10000 times faster than LC-KSVD and 10 times faster than DADL and SK-SVDADL. Note that the distributed ADL (DSADL) used only 510 atoms, but it still achieves a comparable result with the fastest testing time.}

\begin{figure}[htb]
	\centering
	\includegraphics[width=0.45\textwidth]{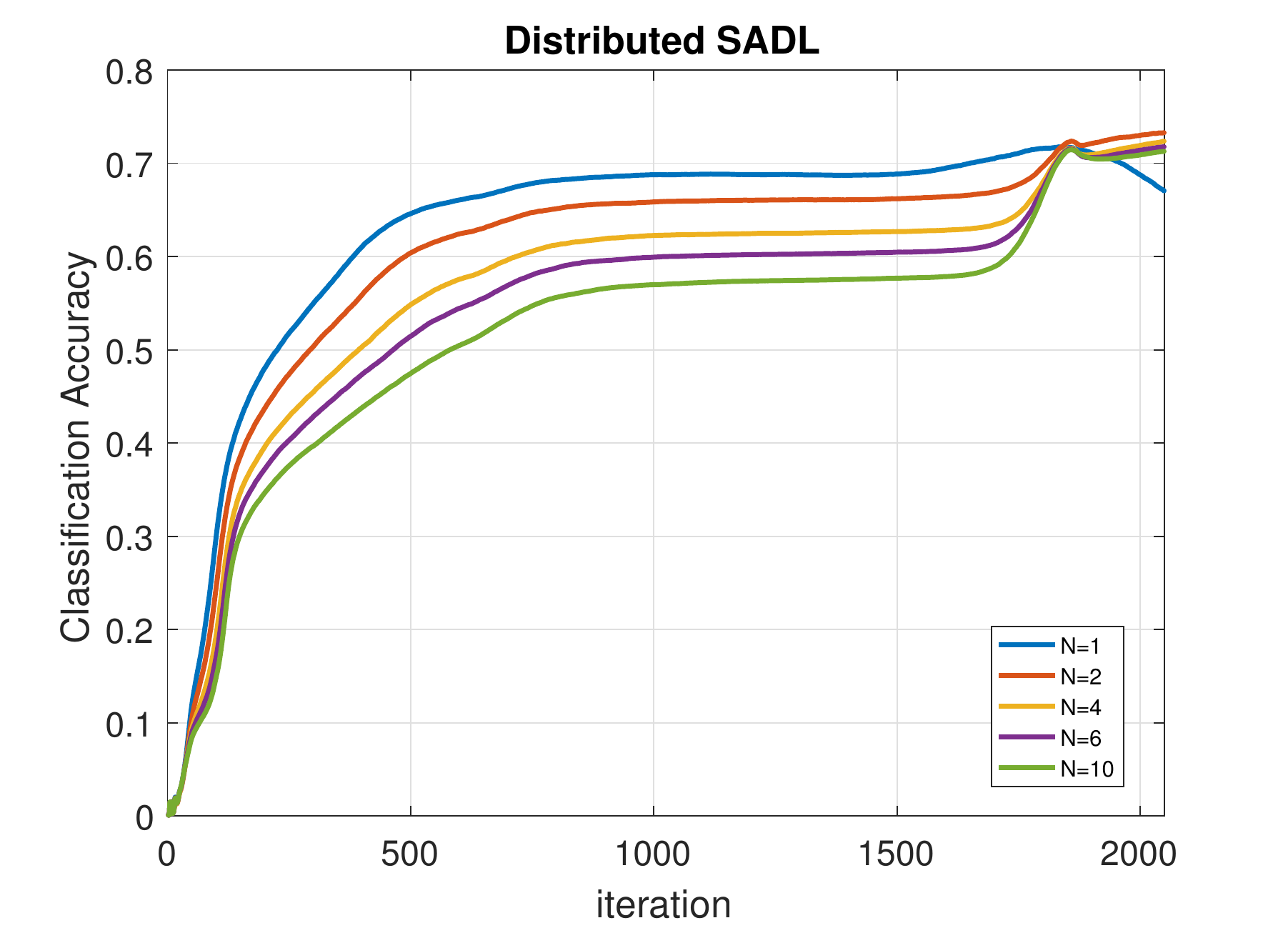}
	\caption{Distributed SADL on Caltech 101: $N$ is the number of clusters used. $N=1$ is centralized. Training set is divided into $N$ groups.}
	\label{fig:distributed}
\end{figure}

The parameters in DSADL are set as the following: $\lambda_1=0.001,\lambda_2=4.6, p=1110$ and the penalty coefficients of the communication cost $\xi_{1_t}=\xi_{2_t}=\xi_{3_t}=0.1, \forall t$. 
Figure \ref{fig:distributed} shows that when the number of groups is increased, the accuracy is actually lower at first because of the smaller training sample size of each independent variable. But after the communication between global variables and local independent variables are enhanced, the performance rises up very quickly to a high generalized accuracy. Distributed SADL is demonstrated that it can also obtain a very stable and excellent performance even when the number of groups is large. 

\begin{figure}[htb]
	\centering
	\includegraphics[width=0.45\textwidth]{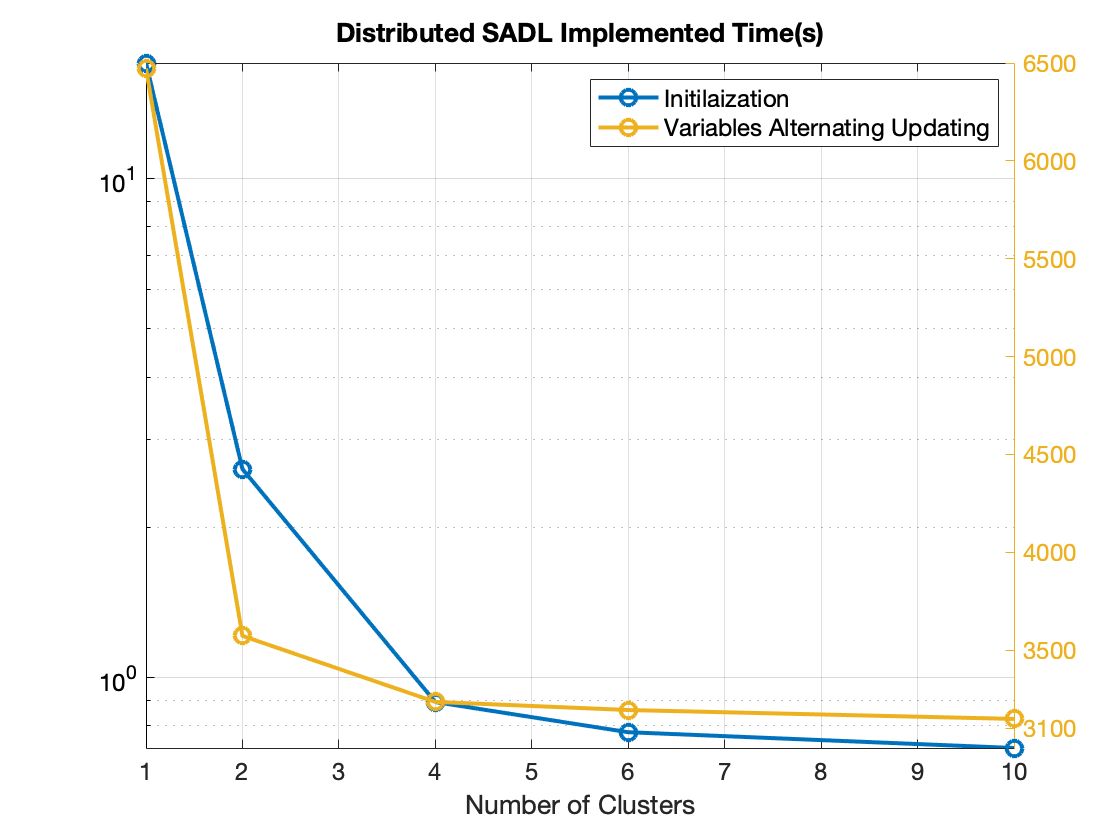}
	\caption{Training time of Distributed SADL on Caltech 101: horizontal axis is the number of cluster, vertical axis is the real training time in second. The left blue vertical axis is the time for initialization of DSADL, and the right orange vertical axis is the time for variables alternating iteration, \emph{i.e.,} while loop part in our Algorithm 2.}
	\label{fig:distributedtime}
\end{figure}

\begin{table}[htb]
    \centering
    \caption{Training Time and \# Training Samples on Caltech 101 Dataset}
    \label{tab:distributedtime}
    \resizebox{\columnwidth}{!}{
    \begin{tabular}{c|c|c|c|c}
        \hline
        &Initiali &Variables  &Total Training  &\#Training Samples\\
        &-zation (s) &Updating (s)&Time (s)&of Each Cluster\\
        \hline
         1 Cluster &17.04  &6471.9  &6488.94  &3060 \\
         2 Clusters& 2.62 & 3572.9 & 3575.52 & 1530\\ 
         4 Clusters& 0.89& 3235.5&3236.39 & 765\\
         6 Clusters& 0.78 &3194.0&3194.78&510\\
         10 Clusters& 0.72&3148.6&3149.32 &306\\
         \hline
    \end{tabular}}
\end{table}

To further study the efficiency of distributed SADL, we conduct an experiment based on different numbers of clusters, which is shown in Fig. \ref{fig:distributedtime}. For fairness, we first utilize only one core in our CPU to run the SADL, while the 2-cluster experiment uses 2 cores to implement DSADL; 4-cluster experiment uses 4 cores on DSADL, and so on. The training time and the number of training samples of each cluster are averaged over 10 realizations and are listed in Table \ref{tab:distributedtime}. It is worth noting that the training time in Table \ref{tab:caltech101} is based on 28 cores (whole cores) in CPU, while the training time in Table \ref{tab:distributedtime} is based on only one core of the CPU. We separate the algorithm of DSADL into two parts: an initialization part and a variable updating part. The initialization part corresponds to the line 1 in Algorithm 2, and the variable updating part is started at line 2 to line 21, \emph{i.e.}, the while loop. The initialization part consists of simple matrix assignments, while the variable updating part has more matrix calculation, such as multiplication and inversion. It is shown in Fig. \ref{fig:distributedtime} that the running time of both the initialization part and the variable updating part quickly decrease when the numbers of clusters increase. The slopes of both curves decrease when more clusters are used, which is due to the fact that the training samples in each cluster is small enough to affect the calculation capability of each CPU core. As there are three global communication terms in Algorithm 2 after updating individual dictionaries, transforming matrix and classifier learnt, the training time with 2 clusters, is slightly more than the half the running time of 1 cluster (centralized). However, these three terms are not expensive, and Algorithm 2, with 2 clusters is still 1.8 times faster than the centralized one. We observed that the more clusters we use, the more training time is saved. Moreover, the larger data is, the more training time is also saved.

\subsection{Caltech 256}
\begin{figure}[htb]
	\centering
	\includegraphics[width=0.45\textwidth]{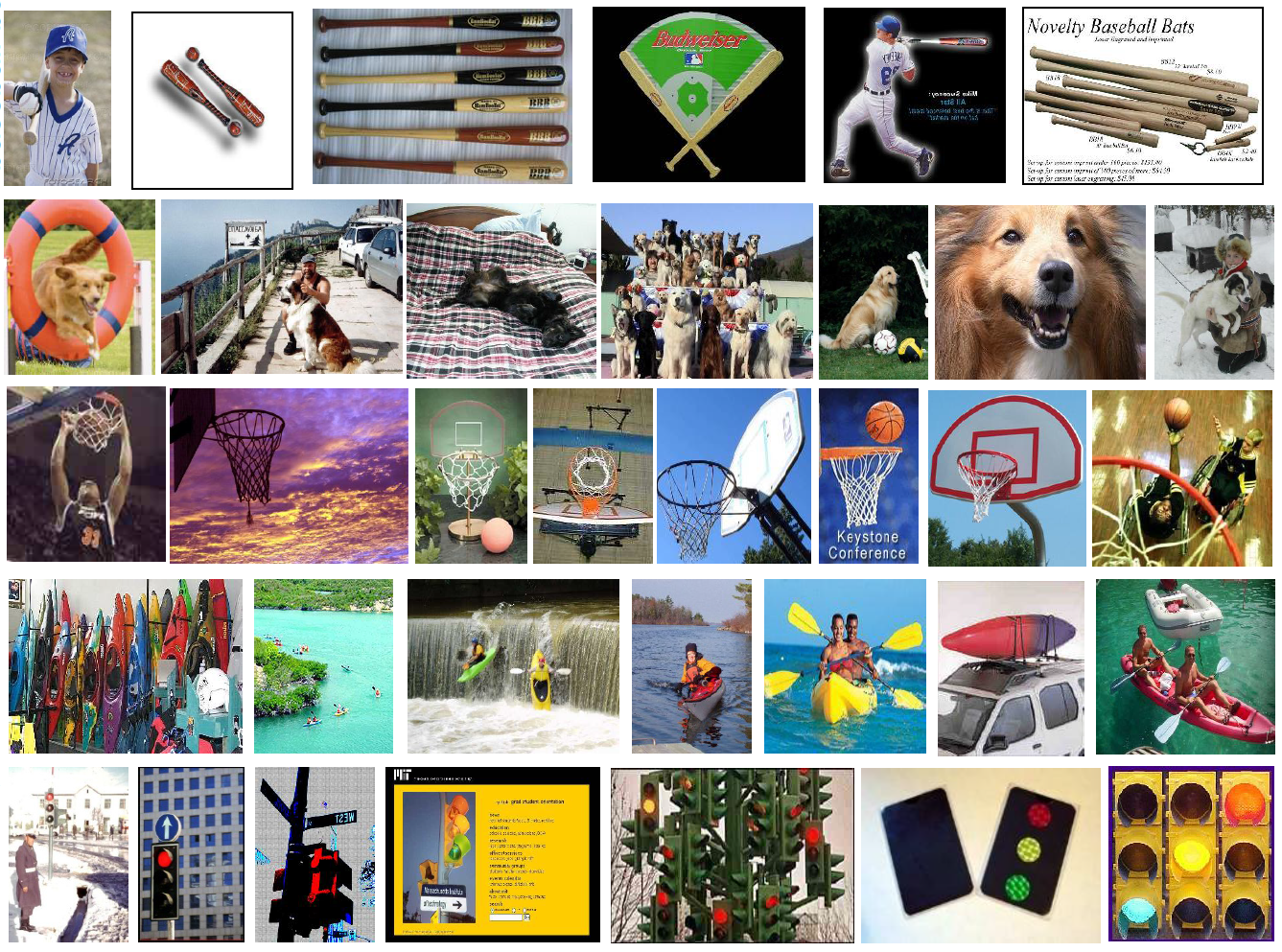}
	\caption{Caltech 256 Dataset Examples}
	\label{fig:caltech256}
\end{figure}

The Caltech 256 is a relatively larger objective dataset, which includes 256 object categories and one clutter. There are totally 30608 images with various object location, pose, and size. Figure \ref{fig:caltech256} shows examples of the Caltech 256 dataset, whose each category has at least 80 images. Note that Caltech 256 includes no rotation or alignment characteristics. Thus, it contains large intra-class diversity and inter-class similarity, such as object scale, object rotation and common patterns. The features of Caltech 256 images are extracted by using the output features of the last layer before fully connected layer of ResNet-50 \cite{he2016deep} with the weights trained by ImageNet. The dimension of each feature is $2046 \times 1$. We randomly sample 15 images from each category for training, and test on the rest of them. To train the Distributed SADL, the dictionary size is set to 3855, dataset is divided into $3$ subsets (\emph{i.e.,} $t=3$ in Algorithm 2), $\lambda_1=0.001, \lambda_2=0.5$, $\xi_{1_t}=\xi_{2_t}=\xi_{3_t}=3 \times 10^{-5}, \forall t$ and $p=4495$.

\begin{table}[htb]
	\centering
	\caption{Classification Results on Caltech 256 Dataset}
	\begin{tabular}{llll}
		\hline
		Methods & Classification & Training & Testing \\
		(training samples) & Accuracy(\%) & Time(s) & Time(s)\\
		\hline
		ADL+SVM(15)\cite{shekhar2014analysis} & 66.66\% &3501.44  & 7.67$\times 10^{-2}$\\
		LC-KSVD(15)\cite{lcksvd} & 73.37\% & 3118.76 & 3.00$\times 10^{-3}$\\
		DADL\cite{dadl} & 72.20\% & 417.06 & 5.42$\times 10^{-4}$\\
		CADL\cite{cadl} & {75.25}\% & 5586.21 & 4.83$\times 10^{-5}$\\
		SK-SVDADL\cite{sksvdadl} & 73.35\% &334.31& 3.28$\times 10^{-4}$\\
		CNN Features(15)\cite{zeiler2014visualizing} &65.70\%\cite{zeiler2014visualizing} &- &- \\
		SADL(15) &\textbf{75.36}\% &4829.01&\textbf{2.79$\times 10^{-5}$}\\
		DSADL(15) &\textbf{74.38}\% &- &\textbf{2.79$\times 10^{-5}$}\\
		\hline
		ResFeats-50(30)\cite{mahmood2016resfeats} &75.40\%\cite{mahmood2016resfeats} & &\\
		
		\hline
	\end{tabular}%
	\label{tab:caltech256}%
\end{table}%

We use Caltech 256 to test both SADL and Distributed SADL. Our SADL achieves the highest accuracy, and our Distributed SADL also achieves a comparable performance with an extremely fast testing time, even though the dimension of the features are increased. For reference, we also compare our method with two network methods \cite{zeiler2014visualizing,mahmood2016resfeats}. In \cite{zeiler2014visualizing}, Zeiler \textit{et al.} constructed a convolutional network per-trained by ImageNet, and then learned an adapted convolutional network for Caltech 256 based on the features of the former network. As trained by 15 samples of each class, our performance is $10\%$ higher than the CNN result. ResFeats-50\cite{mahmood2016resfeats} is a most recent convolutional network method. This method is trained by 30 samples of each category with 50 layers. Though ResFeats-50 utilizes twice more training samples than ours, our result is still very comparable.

\subsection{Scene 15}
\begin{figure}[htb]
	\centering
	\includegraphics[width=0.45\textwidth]{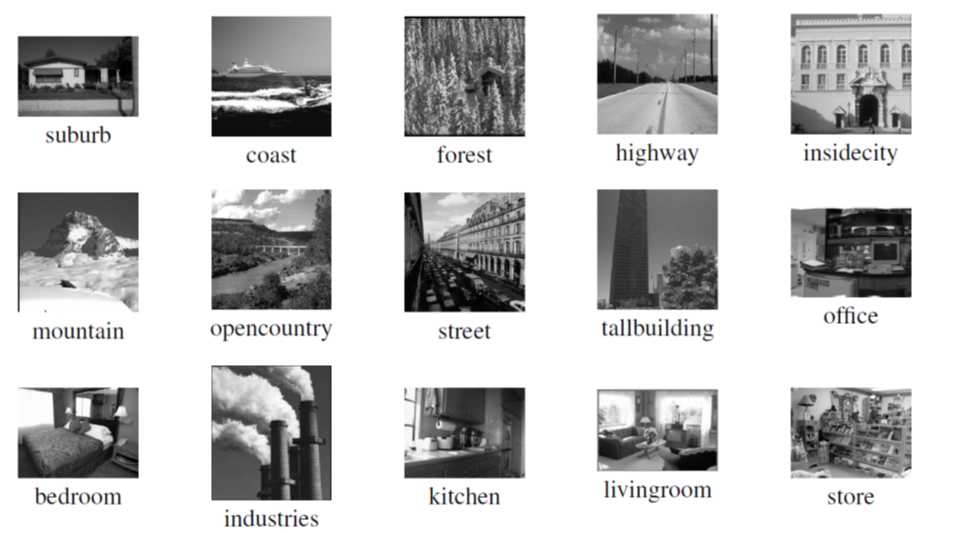}
	\caption{Scene 15 Dataset Examples}
	\label{fig:scene15}
\end{figure}
Scene 15 dataset contains a total of 15 categories of different scenes, and each category has around 200 images. The examples are listed in Figure \ref{fig:scene15}. As different scenes contain many common components, and different components aslo share a large number of common features, training on the Scene 15 dataset is prone to a remarkable amount of inter-class similarity. Proceeding as for the Caltech 101 dataset, we compute the spatial pyramid features for scene images. A four-level spatial pyramid (\emph{i.e.,} each image is grid into $1 \times 1$, $2\times 2$, $4\times 4$ and $8 \times 8$) and a codebook of size 200 is used. The final features are obtained by applying PCA to reduce the dimension of spatial pyramid features to $3000$. We randomly pick 100 images per class as training data, and use the rest of images as testing data. The settings and steps follow \cite{lcksvd}. The dictionary size is set to 1500, $\lambda_1=0.001, \lambda_2=0.003$ and $p=283$.

\begin{table}[htb]
	\centering
	\caption{Classification Results on Scene 15 Dataset}
	\begin{tabular}{llll}
		\hline
		Methods & Classification & Training & Testing \\
		& Accuracy(\%) & Time(s) & Time(s)\\
		\hline
		ADL+SVM\cite{shekhar2014analysis} & 80.55\% & {484.41} & 1.73$\times 10^{-4}$\\
		SRC\cite{src} & 91.80\% & No Need & 4.06$\times 10^{-1}$\\
		LC-KSVD\cite{lcksvd} & \textbf{98.83}\% (92.9\%\cite{lcksvd}) & 390.22 & 1.81$\times 10^{-3}$\\
		DADL\cite{dadl} & 97.81\% (98.3\%\cite{dadl}) & \textbf{33.03} & 4.62 $\times 10^{-4}$\\
		CADL\cite{cadl} & {98.49\% (98.6\%\cite{cadl})} & 4637.80 & 6.02 $\times 10^{-5}$\\
		SK-SVDADL\cite{sksvdadl} & 96.84\% (97.4\%\cite{sksvdadl}) &66.79& 1.06 $\times 10^{-4}$\\
		SADL & 98.50\% & 174.20 &\textbf{2.41$\times 10^{-5}$}\\
		\hline
	\end{tabular}%
	\label{tab:scene15}%
\end{table}%

{The classification results, training and testing time are summarized in Table \ref{tab:scene15}. Our performance is slightly lower than LC-KSVD, but is still higher than all other methods. However, the testing phase is superior to the others. Note that the testing time is the fastest, and the training time is faster than CADL, LC-KSVD and ADL+SVM.}  


\subsection{Comparative Evaluation}
To investigate the effect of different constrains in the optimization problem in Eq. (\ref{equ:ssadl}), we respectively learn ADL by neglecting one of the two constraints and test the resulting algorithms on 10 realizations. The results of 5 datasets are compared with SADL in the Table \ref{tab:constraintscompare}. As there is no linear classifier, training ADL with only the constraint $H=QU$, we assign the class labels $y$ for an observed image $x$ by $y=\min_c\|H_c-Q(Dx)\|_F^2$. For ADL with only the constraint $L=WU$, labels of images are assigned by the classifier $W$.
\begin{table}[htb]
	\centering
	\caption{Classification Results of Different Constraints Comparison}
		\resizebox{\columnwidth}{!}{%
	\begin{tabular}{cccccc}
		\hline
		ADL+constrains& YaleB  & AR & Caltech 101 & Scene 15 & Caltech 256\\
		\hline
		$H=QU$ & 95.37\% & 96.70\% & 74.44\% & 98.46\% & 75.05\%\\
		$L=WU$& 95.52\% & 97.12\% & 74.45\% &98.36\% & 75.35\%\\
		SADL & 96.35\% & 97.13\% &74.45\% &98.50\% & 75.36\%\\
		\hline
	\end{tabular}%
	}
	\label{tab:constraintscompare}%
\end{table}%

The results show that both of these two constraints exhibit a similar behavior, the universal classifier has a slightly better performance, while jointly learning two constraints achieves the best result. The results also support our goal of mitigating the inter-class common features. Therefore, our algorithm achieves a better performance in Scene 15 and Caltech 256, which have many common features among classes.

\section{Conclusion}\label{sec:conclusion1}
We proposed an image classification method referred to as Structured Analysis Dictionary Learning (SADL). To obtain SADL, we constrain a structured subspace (cluster) model in the enhanced ADL method, where each class is represented by a structured subspace. The enhancement of ADL is realized by constraining the learning by a classification fidelity term on the sparse coefficients. Our formulated optimization problem was efficiently solved by the linearized ADM method, in spite of its non-convexity due to bilinearity. Taking advantage of analysis dictionary, our method achieves a significantly faster testing time. Furthermore, a Distributed SADL (DSADL) was also proposed to address the scalability problem. Both discriminative structure and fast testing phase are well preserved in the DSADL. Even though the algorithm was run by many multi-clusters, the performance was still stable and comparable to the centralized SADL. \par

Our experiments demonstrate that our approach has at least a comparable, and often a better performance than state-of-the-art techniques on five well known datasets and achieves superior training and testing times by orders of magnitude.\par

A possible future direction for improving our method could be to leverage the discriminative nature of the synthesis dictionary and the efficiency of the analysis dictionary together. This can achieve a more discriminative power and high efficiency.\par


%

\appendices

\section{Proof of the Algorithm1}
Take the Lagrangian function
\begin{equation}\label{equ:augmentedL}
\begin{split}
&L(\Omega, U, Q, W,\varepsilon^{(1)},\varepsilon^{(2)}, Z^{(1)}, Z^{(2)})=\\
&\frac{1}{2}\|U-\Omega X\|^2_F+\lambda_1\|U\|_1+\frac{\rho_1}{2}\| \varepsilon_1 \|_2^2+\frac{\rho_2}{2}\| \varepsilon_2 \|_2^2\\
&+ \langle Z^{(1)},H-QU-\varepsilon_1 \rangle + \langle Z^{(2)},Y-WQU-\varepsilon_2 \rangle\\
&+\frac{\mu}{2} \|H-QU-\varepsilon_1 \|_2^2+ \frac{\mu}{2} \|Y-WQU-\varepsilon_2\|_2^2\\
&+\frac{\delta_1}{2}\|Q\|_2^2+\frac{\delta_2}{2}\| W \|_2^2+\frac{\lambda_2}{2}\| \Omega \|_2^2.\\
\end{split}
\end{equation}
Our algorithm can be written as the one in Alg. \ref{alg:learning}. 
\begin{algorithm}
	\caption{Linearized ADM for Structured Analysis Dictionary Learning}
	\label{alg:learning}
	At each iteration $k+1$, compute:
	\begin{equation}\label{equ:Uupdate}
	\resizebox{0.88\columnwidth}{!}{$U_{k+1}=\tau_{\frac{\lambda_1}{\mu\eta_U}}\left(U_k-\frac{1}{\mu\eta_U}\bigtriangledown_UL_s(U_k,Q_k,W_k,\Omega_k,\varepsilon_k^{(1)},\varepsilon_k^{(2)},Z_k^{(1)},Z_k^{(2)})\right),$}
	\end{equation}
	\begin{equation}\label{equ:Qupdate}
	\resizebox{0.88\columnwidth}{!}{$Q_{k+1}=Q_k-\frac{1}{\mu\eta_Q}\bigtriangledown_QL(U_{k+1},Q_k,W_k,\Omega_k,\varepsilon_k^{(1)},\varepsilon_k^{(2)},Z_k^{(1)},Z_k^{(2)}),$}
	\end{equation}
	\begin{equation}\label{equ:Wupdate}
	\resizebox{0.88\columnwidth}{!}{$W_{k+1}=W_k-\frac{1}{\mu\eta_W}\bigtriangledown_WL(U_{k+1},Q_{k+1},W_k,\Omega_k,\varepsilon_k^{(1)},\varepsilon_k^{(2)},Z_k^{(1)},Z_k^{(2)}),$}
	\end{equation}
	\begin{equation}\label{equ:Dupdate}
	\resizebox{0.88\columnwidth}{!}{$\Omega_{k+1}=\arg\min_\Omega L(U_{k+1},Q_{k+1},W_{k+1},\Omega,\varepsilon_k^{(1)},\varepsilon_k^{(2)},Z_k^{(1)},Z_k^{(2)}),$}
	\end{equation}
	\begin{equation}\label{equ:eps1update}
	\resizebox{0.88\columnwidth}{!}{$\varepsilon^{(1)}_{k+1}=\arg\min_{\varepsilon^{(1)}} L(U_{k+1},Q_{k+1},W_{k+1},\Omega_{k+1},\varepsilon^{(1)},\varepsilon_k^{(2)},Z_k^{(1)},Z_k^{(2)}),$}
	\end{equation}
	\begin{equation}\label{equ:eps2update}
	\resizebox{0.88\columnwidth}{!}{$\varepsilon^{(2)}_{k+1}=\arg\min_{\varepsilon^{(2)}} L(U_{k+1},Q_{k+1},W_{k+1},\Omega_{k+1},\varepsilon_{k+1}^{(1)},\varepsilon^{(2)},Z_k^{(1)},Z_k^{(2)}),$}
	\end{equation}
	\begin{equation}\label{equ:Y1update}
	\scalebox{0.73}{$Z_{k+1}^{(1)}=Z_{k}^{(1)}+\mu(H-Q_{k+1}U_{k+1}-\varepsilon^{(1)}_{k+1}),$}
	\end{equation}
	\begin{equation}\label{equ:Y2update}
	\scalebox{0.73}{$Z_{k+1}^{(2)}=Z_{k}^{(2)}+\mu(Y-W_{k+1}Q_{k+1}U_{k+1}-\varepsilon^{(2)}_{k+1}),$}
	\end{equation}
\end{algorithm}
Let us proceed by introducing two simple lemmas:
\begin{lemma}\label{lemma:bound_lip}
	Consider a differentiable function $f$ with an $L-$Lipschitz continuous derivative and another arbitrary convex function $g$. For any arbitrary point $x$ define
	\[
	x^+=\mathrm{prox}_{\tau g}(x-\tau\nabla f(x)),
	\]
	where $\tau>0$ is a step size and
	\[
	\mathrm{prox}_{\tau g}(y)=\arg\min\limits_{x}\frac 1 2\|x-y\|^2+\tau g(x).
	\]
	Then, we have \[F(x^+)-F(x)\leq \left(\frac L 2-\frac 1\tau\right)\|x-x^+\|^2,\] where $F(x)=f(x)+g(x)$.
	\begin{proof}
		Notice that by the definition of the proximal operator $\mathrm{prox}$, there exists a subgradient $\xi\in\partial g(x^+)$ such that
		\[
		x^+=x^--\tau\xi,
		\]
		where $x^-=x-\tau\nabla f(x)$. Then, we have
		\[
		g(x)\geq g(x^+)+\langle x-x^0,\xi\rangle.
		\]
		On the other hand,
		\[
		f(x)\geq f(x^+)+\langle x-x^+,\nabla f(x)\rangle-\frac L 2\|x-x^+\|^2.
		\]
		Adding the two inequalities yields
		\[
		F(x)\geq F(x^+)+\langle x-x^+,\nabla f(x)+\xi\rangle-\frac L 2\|x-x^+\|^2.
		\]
		Now noticing that $\tau(\nabla f(x)+\xi)=x-x^+$ completes the proof.
	\end{proof}
\end{lemma}

\begin{lemma}\label{lemma:stat}
	Consider a differentiable function $f$ and a convex function $g$. Suppose that a point $x$ satisfies
	\[
	\mathrm{prox}_{\tau g}(x-\tau\nabla f(x))=x.
	\]
	Then, $x$ is a stationary point of $F=f+g$, i.e. $-\nabla f(x)\in\partial g(x)$.
	\begin{proof}
		From the definition of the proximal operator there exists a vector $\xi\in \partial g(x)$ such that $x=x-\tau\nabla f(x)-\tau\xi$. We conclude that $-\nabla f(x)=\xi$, which completes the proof.
	\end{proof} 
\end{lemma}

Next, we make a simple but crucial observation about our algorithm:
\begin{lemma}\label{lemma:dualboundedbypriaml}
	For Algorithm \ref{alg:learning} the following holds in every iteration $k$:
	\[
	Z^{(1)}_{k+1}=\rho_1 \varepsilon^{(1)}_{k+1},
	\]
	\[
	Z^{(2)}_{k+1}=\rho_2 \varepsilon^{(2)}_{k+1},
	\]
	and as a result,
	\begin{equation}
	\|Z_{k+1}^{(1)}-Z_{k}^{(1)}\| = \rho_1\|\varepsilon_{k+1}^{(1)}-\varepsilon_{k}^{(1)}\|,
	\end{equation}
	\begin{equation}
	\|Z_{k+1}^{(2)}-Z_{k}^{(2)}\| = \rho_2\|\varepsilon_{k+1}^{(2)}-\varepsilon_{k}^{(2)}\|.
	\end{equation}
	\begin{proof}
		From the $\varepsilon^{(1)}$ update rule (\ref{equ:eps1update}),we have the following optimality condition
		\begin{equation}\label{equ:eps1argmin}
		\rho_1 \varepsilon^{(1)}_{k+1}-Z^{(1)}_{k}-\mu(H-Q_{k+1}U_{k+1}-\varepsilon^{(1)}_{k+1})=0.
		\end{equation}
		Combining with dual variable $Z^{(1)}_{k+1}$ update rule (\ref{equ:Y1update}), we obtain
		\begin{equation}\label{equ:y1eps1}
		Z^{(1)}_{k+1}=\rho_1 \varepsilon^{(1)}_{k+1}.
		\end{equation}
		The result for $Z^{(2)}_{k+1}$ is similarly obtained.
	\end{proof}
\end{lemma}
We take $L_k=L(\Omega_k,U_k,Q_k,W_k,\varepsilon^{(1)}_k,\varepsilon^{(2)}_k,Z^{(1)}_k,Z^{(2)}_k,\mu_k)$ for $k=0,1,2,\ldots$ and notice that the change in $L_k$ can be controlled by the following result:
\begin{lemma}\label{lemma:iter}
	\begin{equation}
	\begin{split}
	&L_{k+1}-L_k\\
	&\resizebox{0.88\columnwidth}{!}{$\leq \left(\frac{\alpha_{k,U}}{2}-\mu\eta_U\right) \|U_{k+1}-U_k\|^2+\left(\frac{\alpha_{k,Q}}{2}-\mu\eta_Q\right) \|Q_{k+1}-Q_k\|^2$}\\
	&\resizebox{0.88\columnwidth}{!}{$+\left( \frac{\alpha_{k,W}}{2}-\mu\eta_W\right) \|W_{k+1}-W_k\|^2-\frac{m_\Omega}{2}\|\Omega_{k+1}-\Omega_{k}\|^2$}\\
	&\resizebox{0.88\columnwidth}{!}{$+\left(\frac{\rho_1^2}{\mu}-\frac{m_{\varepsilon^{(1)}}}{2}\right)\|\varepsilon^{(1)}_{k+1}-\varepsilon^{(1)}_{k}\|^2
		+\left(\frac{\rho_2^2}{\mu}-\frac{m_{\varepsilon^{(2)}}}{2}\right)\|\varepsilon^{(2)}_{k+1}-\varepsilon^{(2)}_{k}\|^2,$}\\
	\end{split}
	\end{equation}
	where
	\[
	\alpha_{k,U}=1+\mu\|Q_k^TQ_k+Q_k^TW_k^TW_kQ_k\|_*,
	\]
	\[
	\alpha_{k,Q}=\delta_1+\mu\|W_k^TW_k\|_*\|U_{k+1}U_{k+1}^T\|_*,
	\]
	\[
	\alpha_{k,W}=\delta_2+\mu\|Q_{k+1}U_{k+1}U^T_{k+1}Q^T_{k+1}\|_*,
	\]
	\[
	m_\Omega=\sigma_{\mathrm{min}}(XX^T)
	\]
	\[
	m_{\varepsilon^{(1)}}=\rho_1+\mu,\quad m_{\varepsilon^{(2)}}=\rho_2+\mu .
	\]
	\begin{proof}
		Respectively denote by $\Delta L_{k,U},\Delta L_{k,Q},\Delta L_{k,W}, \Delta L_{k,\Omega},\Delta L_{k,\varepsilon^{(i)}}, \Delta L_{k,Y^{(i)}}$ for $i=1,2$, the change in $L$ corresponding to the update of $U,Q,W,\Omega,\varepsilon^{(i)}$ and $Y^{(i)}$ in Eq. (2-9). Notice that
		\[
		L_{k+1}-L_k=\Delta L_{k,U}+\Delta L_{k,Q}+\Delta L_{k,W}+ \Delta L_{k,\Omega}
		\]
		\[+\Delta L_{k,\varepsilon^{(1})}+\Delta L_{k,\varepsilon^{(2})}+ \Delta L_{k,Z^{(1)}} +\Delta L_{k,Z^{(2)}}.
		\]
		Notice that by taking $f(U)=L_s(\Omega_k,U,Q_k,W_k,\varepsilon^{(1)}_k,\varepsilon^{(2)}_k,Z^{(1)}_k,Z^{(2)}_k)$, $g(U)=\lambda_1\|U\|_1$ and $\tau=1/\mu\eta_U$, and recalling Lemma \ref{lemma:bound_lip}, we have
		\begin{equation}\label{eq:boundU}
		\Delta L_{k,U}\leq\left(\frac{\alpha_{k,U}}{2}-\mu\eta_U\right)\left\|U_{k+1}-U_k\right\|^2,
		\end{equation}
		where we use the fact that $f(U)$ is quadratic, hence possessing $\alpha_{k,U}-$Lipschitz derivatives with $\alpha_{k,U}$ being the largest singular value of the Hessian. Similarly, by taking $f(Q)=L(\Omega_k,U_{k+1},Q,W_k,\varepsilon^{(1)}_k,\varepsilon^{(2)}_k,Z^{(1)}_k,Z^{(2)}_k), g(Q)=0, \tau=1/\mu\eta_Q$ and $f(W)=L(\Omega_k,U_{k+1},Q_{k+1},W,\varepsilon^{(1)}_k,\varepsilon^{(2)}_k,Z^{(1)}_k,Z^{(2)}_k), g(W)=0, \tau=1/\mu\eta_U$ and utilizing Lemma \ref{lemma:bound_lip}, we respectively obtain
		\begin{equation}\label{eq:boundQ}
		\Delta L_{k,Q}\leq\left(\frac{\alpha_{k,Q}}{2}-\mu\eta_Q\right)\left\|Q_{k+1}-Q_k\right\|^2,
		\end{equation}
		\begin{equation}\label{eq:boundW}
		\Delta L_{k,W}\leq\left(\frac{\alpha_{k,W}}{2}-\mu\eta_W\right)\left\|W_{k+1}-W_k\right\|^2.
		\end{equation}
		Next, notice that the function $f(\Omega)=L(\Omega,U_{k+1},Q_{k+1},W_{k+1},\varepsilon^{(1)}_k,\varepsilon^{(2)}_k,Z^{(1)}_k,Z^{(2)}_k)$ is quadratic and $m_\Omega$ is strongly convex, where $m_\Omega$ is the smallest singular value of Hessian. Hence,
		\begin{equation}\label{eq:boundOmega}
		\Delta L_{k,\Omega}=f(\Omega_k)-\min_\Omega f(\Omega)\leq-\frac{m_\Omega}{2}\|\Omega_{k+1}-\Omega_{k}\|^2
		\end{equation}
		Similarly, taking $f(\varepsilon^{(1)})=L(\Omega_{k+1},U_{k+1},Q_{k+1},W_{k+1},\varepsilon^{(1)},\varepsilon^{(2)}_k,Z^{(1)}_k,Z^{(2)}_k)$ and $f(\varepsilon^{(2)})=L(\Omega_{k+1},U_{k+1},Q_{k+1},W_{k+1},\varepsilon^{(1)}_{k+1},\varepsilon^{(2)},Z^{(1)}_k,Z^{(2)}_k)$, we obtain
		\begin{equation}\label{eq:boundEpsilon}
		\Delta L_{k,\varepsilon^{(i)}}\leq-\frac{m_{\varepsilon^{(i)}}}{2}\left\|\varepsilon^{(i)}_{k+1}-\varepsilon^{(i)}_{k}\right\|^2,\quad i=1,2.
		\end{equation}
		Finally, notice that
		\begin{equation*}\label{eq:boundY1}
		\begin{split}
		&\Delta L_{k,Z^{(1)}}=\left\langle Z^{(1)}_{k+1}-Z^{(1)}_{k}, H-Q_{k+1}U_{k+1}-\varepsilon^{(1)}_{k+1}\right\rangle
		\nonumber\\
		&=\left\langle Z^{(1)}_{k+1}-Z^{(1)}_{k}, \frac 1\mu\left(Z^{(1)}_{k+1}-Z^{(1)}_{k}\right)\right\rangle
		=\frac 1\mu\left\|Z^{(1)}_{k+1}-Z^{(1)}_{k}\right\|^2\\
		&=\frac {\rho_1^2}\mu\left\|\varepsilon^{(1)}_{k+1}-\varepsilon^{(1)}_{k}\right\|^2.
		\end{split}
		\end{equation*}
		Similarly, we obtain
		\begin{equation}\label{eq:boundY2}
		\Delta L_{k,Z^{(2)}}=\frac {\rho_2^2}\mu\left\|\varepsilon^{(2)}_{k+1}-\varepsilon^{(2)}_{k}\right\|^2.
		\end{equation}
		Summing the inequalities in Eq.~\eqref{eq:boundU}, Eq.~\eqref{eq:boundQ}, Eq.~\eqref{eq:boundW}, Eq.~\eqref{eq:boundOmega}, Eq.~\eqref{eq:boundEpsilon}, Eq.~\eqref{eq:boundY1} and Eq.~\eqref{eq:boundY2} completes the proof.
	\end{proof}
\end{lemma}
Now, we have the following theorem:
\begin{theorem}\label{theorem:main}
	Suppose that $\mu\geq \sqrt 2\{\rho_1,\rho_2\}$. There exist positive values $\eta^0_U,\eta^0_Q,\eta^0_W$ only depending on the initial values such that for $\eta_U>\eta^0_U, \eta_Q>\eta^0_Q, \eta_W>\eta^0_W$ the sequence $\{L_k\}_{k=1}^\infty$ is positive and decreasing, hence convergent.
	\begin{proof}
		First define
		\[
		L_{k,e}(\Omega, U, Q, W)=L(\Omega, U, Q, W, \varepsilon_k^{(1)},\varepsilon_k^{(2)}, Z_k^{(1)}, Z_k^{(2)}).
		\]
		According to Lemma \ref{lemma:dualboundedbypriaml}, for $k=1,2,\ldots$, we have
		\begin{equation}\label{eq:Le}
		\begin{split}
		&L_{k,e}=\frac{1}{2}\|U-\Omega X\|^2_F+\lambda_1\|U\|_1\\
		&\resizebox{1\columnwidth}{!}{$+ \rho_1\langle \varepsilon^{(1)}_k,H-QU-\varepsilon^{(1)}_k \rangle + \frac{\mu}{2} \|H-QU-\varepsilon^{(1)}_k \|_2^2+\frac{\rho_1}{2}\| \varepsilon^{(1)}_k \|_2^2$}\\
		&\resizebox{1\columnwidth}{!}{$+\rho_2\langle \varepsilon^{(2)}_k,Y-WQU-\varepsilon^{(2)}_k \rangle+ \frac{\mu}{2} \|Y-WQU-\varepsilon^{(2)}_k\|_2^2+\frac{\rho_2}{2}\| \varepsilon^{(2)}_k \|_2^2$}\\
		&+\frac{\delta_1}{2}\|Q\|_2^2+\frac{\delta_2}{2}\| W \|_2^2+\frac{\lambda_2}{2}\| \Omega \|_2^2\\
		&=\frac{1}{2}\|U-\Omega X\|^2_F+\lambda_1\|U\|_1\\
		&+ \frac{\mu}{2} \left\|H-QU-\left(1-\frac{\rho_1}{\mu}\right)\varepsilon^{(1)}_k \right\|_2^2
		+\frac{\rho_1}{2}\left(1-\frac{\rho_1}{\mu}\right)\| \varepsilon^{(1)}_k \|_2^2\\
		&+ \frac{\mu}{2} \left\|Y-WQU-\left(1-\frac{\rho_2}{\mu}\right)\varepsilon^{(2)}_k\right\|_2^2+\frac{\rho_2}{2}\left(1-\frac{\rho_2}{\mu}\right)\| \varepsilon^{(2)}_k \|_2^2\\
		&+\frac{\delta_1}{2}\|Q\|_2^2+\frac{\delta_2}{2}\| W \|_2^2+\frac{\lambda_2}{2}\| \Omega \|_2^2.\\
		\end{split}
		\end{equation}
		Hence, we have $L_{k,e}\geq 0$ for $\mu>\max\{\rho_1,\rho_2\}$. In particular, we obtain that $L_k=L_{k,e}(\Omega_k, U_k, Q_k, W_k)\geq 0$. Now, we use complete (strong) indiction to show that $L_{k+1}\geq L_k$ for $k=1,2,\ldots$. Suppose that this holds for $k=1,2,\ldots,t$. We conclude that $L_t\leq L_1$. Now, notice that from \eqref{eq:Le} and the fact that $L_t=L_{t,e}(\Omega_t, U_t, Q_t, W_t)$ we obtain for $\mu>\max\{\rho_1,\rho_2\}$ that
		\[
		\|Q_t\|^2\leq\frac {2L_1}{\delta_1}, \quad \|W_t\|^2\leq\frac {2L_1}{\delta_2},
		\]
		which leads to the following:
		\[
		\alpha_{t,U}\leq 1+\mu\left(\|Q_t\|^2+\|Q_t\|^2\|W_t\|^2\right)\leq 1+\frac{2L_1\mu}{\delta_1}\left(1+\frac{2L_1}{\delta_2}\right).
		\]
		According to \eqref{eq:boundU}, by selecting $\eta_U>\left[2+\frac{2L_1\mu}{\delta_1}\left(1+\frac{L_1}{\delta_2}\right)\right]/2\mu$, we have that
		\begin{equation}\label{eq:boundUr}
		\Delta L_{t,U}\leq-\frac 1 2\left\|U_{t+1}-U_t\right\|^2,
		\end{equation}
		which subsequently yields,
		\[
		L_{t,e}(\Omega_t, U_{t+1}, Q_t, W_t)\leq L_t\leq L_1.
		\]
		Then according to \eqref{eq:Le} for $\mu>\max\{\rho_1,\rho_2\}$, we have that
		\[
		\|U_{t+1}\|_1\leq \frac{L_1}{\lambda_1}.
		\]
		We conclude that
		\[
		\alpha_{t,Q}\leq\delta_1+\mu\|W_t\|^2\|U_{t+1}\|_1^2\leq\delta_1+\mu\frac{2L^2_1}{\lambda_1\delta_2}.
		\]
		Now, by taking $\eta_U>\left[1+\delta_1+\frac{2\mu L^2_1}{\lambda_1\delta_2}\right]/2\mu$ in \eqref{eq:boundQ} we have that
		\begin{equation}\label{eq:boundQr}
		\Delta L_{t,Q}\leq-\frac 1 2\left\|Q_{t+1}-Q_t\right\|^2.
		\end{equation}
		This also results in
		\[
		L_{t,e}(\Omega_t, U_{t+1}, Q_{t+1}, W_t)\leq L_{t,e}(\Omega_t, U_{t+1}, Q_t, W_t)\leq L_t\leq L_1,
		\]
		which using  \eqref{eq:Le} for $\mu>\max\{\rho_1,\rho_2\}$ leads to
		\[
		\|Q_{t+1}\|^2\leq\frac {2L_1}{\delta_1},
		\]
		and hence
		\[
		\alpha_{t,W}\leq\delta_2+\mu\|Q_{t+1}\|^2\|U_{t+1}\|_1^2\leq\delta_2+\frac{2\mu L^2_1}{\delta_1\lambda_1}.
		\]
		Now, by choosing $\eta_W\geq \left[1+\delta_2+\frac{2\mu L^2_1}{\lambda_1\delta_1}\right]/2\mu$ we conclude from \eqref{eq:boundW} that
		\begin{equation}\label{eq:boundWr}
		\Delta L_{t,W}\leq-\frac 1 2\left\|W_{t+1}-W_t\right\|^2.
		\end{equation}
		
		Finally, by choosing $\mu>\sqrt 2\max\{\rho_1,\rho_2\}$, we obtain from Lemma \ref{lemma:iter} that
		\begin{eqnarray}\label{eq:boundR}
		&\resizebox{0.85\columnwidth}{!}{$L_{t+1}-L_t\leq-\frac 1 2\|U_{t+1}-U_t\|_2^2-\frac 1 2\|Q_{t+1}-Q_t\|_2^2-\frac 1 2\|W_{t+1}-W_t\|_2^2 \nonumber$}\\
		&\resizebox{0.85\columnwidth}{!}{$-\frac {m_\Omega} 2\|\Omega_{t+1}-\Omega_t\|_2^2-\frac {\rho_1} 2\|\varepsilon^{(1)}_{t+1}-\varepsilon^{(1)}_t\|_2^2-\frac {\rho_2} 2\|\varepsilon^{(2)}_{t+1}-\varepsilon^{(2)}_t\|_2^2$.}
		\end{eqnarray}
		We conclude that $L_{t+1}\leq L_t$ which completes the proof.
	\end{proof}
\end{theorem}
We finally obtain the following corollary which clarifies the statement and gives the proof of our main result in Theorem \ref{theorem:main}:
\newtheorem{cor}{Corollary}
\begin{cor}
	Suppose that $\mu\geq\sqrt 2\{\rho_1,\rho_2\}$. There exist positive values $\eta^0_U,\eta^0_Q,\eta^0_W, R$ only depending on the initialization such that for $\eta_U>\eta^0_U, \eta_Q>\eta^0_Q, \eta_W>\eta^0_W$ the sequence $\{\Theta_k=(\Omega_k,U_k,Q_k,W_k,\varepsilon_k^{(1)},\varepsilon_k^{(2)}, Z_k^{(1)}, Z_k^{(2)})\}_{k=1}^\infty$ satisfies the following:
	\begin{enumerate}
		\item The parameters for $k=0,1,2,...$ are bounded by $R$, i.e 
		\begin{equation*}
		\resizebox{0.86\columnwidth}{!}{$\|\Theta_k\|=\max\left\{\|\Omega_k\|,\|U_k\|,\|Q_k\|,\|W_k\|,\|\varepsilon_k^{(1)}\|,\|\varepsilon_k^{(2)}\|, \|Z_k^{(1)}\|, \|Z_k^{(2)}\|\right\}<R$}.
		\end{equation*} 
		Hence, they are confined in a compact set.
		\item Any convergence subsequence of $\{\Theta_k\}$ converges to a point $\Theta^*\in S$.
		\item $\mathrm{dist}(\Theta_k,S)$ converges to zero, where
		\[
		\mathrm{dist}(\Theta,S)=\min_{\Theta^\prime\in S}\|\Theta^\prime-\Theta\|.
		\]
	\end{enumerate}
	\begin{proof}
		1) is simply obtained by noticing \eqref{eq:Le} and the fact that $L_{k,e}(\Omega_k,U_k,Q_k,W_k)=L_k\leq L_1$, since $\{L_k\}$ is decreasing. For 2), note that since the sequence $\{L_k\}$ is convergent, we have $\lim_{k\to\infty}L_{k+1}-L_k=0$, which according to \eqref{eq:boundR} yields
		\begin{equation*}
		\begin{split}
		&		\resizebox{1\columnwidth}{!}{$\lim_{k\to\infty}\|U_{k+1}-U_k\|_2^2=\lim_{k\to\infty}\|Q_{k+1}-Q_k\|_2^2
			=\lim_{k\to\infty}\|W_{k+1}-W_k\|_2^2$}\\
		& \resizebox{0.65\columnwidth}{!}{$=\lim_{k\to\infty}\|\Omega_{k+1}-\Omega_k\|_2^2
			=\|\varepsilon^{(i)}_{k+1}-\varepsilon^{(i)}_k\|_2^2=0$,}\\
		\end{split}
		\end{equation*}
		
		\noindent for $i=1,2$. Also from Lemma \ref{lemma:dualboundedbypriaml} we have that
		\[
		\lim_{k\to\infty}\|Z^{(i)}_{k+1}-Z^{(i)}_k\|_2^2=0.
		\]
		We conclude that 
		\begin{equation*}
		\resizebox{0.97\columnwidth}{!}{$\lim_{k\to\infty}\left\|\tau_{\frac{\lambda_1}{\mu\eta_U}}\left(U_k-\frac{1}{\mu\eta_U}\nabla_UL_s(U_k,Q_k,W_k,\Omega_k,\varepsilon_k^{(1)},\varepsilon_k^{(2)},Z_k^{(1)},Z_k^{(2)})\right)-U_k\right\|_2^2
			=0$,}
		\end{equation*}
		\[
		\lim_{k\to\infty}
		\left\|
		\nabla_Q
		L(U_{k+1},Q_k,W_k,\Omega_k,\varepsilon_k^{(1)},\varepsilon_k^{(2)},Z_k^{(1)},Z_k^{(2)})\right\|_2^2
		=0,
		\]
		\[
		\lim_{k\to\infty}
		\left\|
		\nabla_W
		L(U_{k+1},Q_{k+1},W_k,\Omega_k,\varepsilon_k^{(1)},\varepsilon_k^{(2)},Z_k^{(1)},Z_k^{(2)})\right\|_2^2
		=0,
		\]
		\[
		\lim_{k\to\infty}
		\left\|
		H-Q_{k+1}U_{k+1}-\varepsilon^{(1)}_{k+1}\right\|_2^2
		=0,
		\]
		\[
		\lim_{k\to\infty}
		\left\|
		Y-W_{k+1}Q_{k+1}U_{k+1}-\varepsilon^{(2)}_{k+1}\right\|_2^2
		=0.
		\]
		Moreover, note that the Lagrangian $L$ is $L_\Omega-$second order Lipschitz with respect to $\Omega$ (fixing the rest) with $L_\Omega=\|XX^T\|_*$. We obtain that
		\begin{equation*}
		\begin{split}
		&\left\|
		\nabla_\Omega
		L(U_{k+1},Q_{k+1},W_{k+1},\Omega_k,\varepsilon_k^{(1)},\varepsilon_k^{(2)},Z_k^{(1)},Z_k^{(2)})\right\|_2^2\\
		&\leq L_\Omega^2\|\Omega_{k+1}-\Omega_k\|_2^2,\\
		\end{split}
		\end{equation*}
		which yields
		\begin{equation*}
		\resizebox{1\columnwidth}{!}{$\lim_{k\to\infty}
			\left\|
			\nabla_\Omega
			L(U_{k+1},Q_{k+1},W_{k+1},\Omega_k,\varepsilon_k^{(1)},\varepsilon_k^{(2)},Z_k^{(1)},Z_k^{(2)})\right\|_2^2
			=0$.}
		\end{equation*}
		Similarly, we obtain 
		\begin{equation*}
		\resizebox{1\columnwidth}{!}{$\lim_{k\to\infty}
			\left\|
			\nabla_{\varepsilon^{(1)}}
			L(U_{k+1},Q_{k+1},W_{k+1},\Omega_{k+1},\varepsilon_k^{(1)},\varepsilon_k^{(2)},Z_k^{(1)},Z_k^{(2)})\right\|_2^2
			=0$,}
		\end{equation*}
		\begin{equation*}
		\resizebox{1\columnwidth}{!}{$\lim_{k\to\infty}
			\left\|
			\nabla_{\varepsilon^{(2)}}
			L(U_{k+1},Q_{k+1},W_{k+1},\Omega_{k+1},\varepsilon_{k+1}^{(1)},\varepsilon_k^{(2)},Z_k^{(1)},Z_k^{(2)})\right\|_2^2
			=0$.}
		\end{equation*}
		Now, consider a subsequence of $\{\Theta_k\}$ converging to a point $\Theta_*=(\Omega_*,U_*,Q_*,W_*,\varepsilon_*^{(1)},\varepsilon_*^{(2)}, Z_*^{(1)}, Z_*^{(2)})$. Since the argument of the above limits are continuous, we obtain
		\[
		\tau_{\frac{\lambda_1}{\mu\eta_U}}\left(U_*-\frac{1}{\mu\eta_U}\nabla_UL_s(\Theta_*)\right)-U_*=0,
		\]
		\[
		\nabla_Q
		L(\Theta_*)=0,\quad \nabla_W
		L(\Theta_*)=0,\quad  \nabla_{\varepsilon^{(i)}}
		L(\Theta_*)=0,
		\]
		\[
		\nabla_{Z^{(1)}}
		L(\Theta_*)=
		H-Q_*U_*-\varepsilon^{(1)}_*=0,
		\]
		\[
		\quad
		\nabla_{Z^{(2)}}
		L(\Theta_*)=Y-W_*Q_*U_*-\varepsilon^{(2)}_*.
		\]
		According to Lemma \ref{lemma:stat}, we conclude that $\Theta_*\in S$. For 3), suppose that the claim is not true. Then, according to 1) there exists a convergent subsequence of $\{\Theta_k\}$ which is $\gamma-$distant from $S$, \textit{i.e.,} $ \mathrm{dist}(\Theta_k,S)=\gamma>0 $. Then, the convergence point is also $\gamma-$distant from $S$ which contradicts 2) and completes the proof.
		
	\end{proof}
\end{cor}

\section*{Acknowledgment}
We gratefully acknowledge the generous support of the U.S. Army Research Office under grant W911NF-16-2-0005.

\ifCLASSOPTIONcaptionsoff
  \newpage
\fi



%


\bibliographystyle{IEEEtran}
\bibliography{IEEEabrv,egbib}

\end{document}